\documentclass[times,twocolumn,final,authoryear]{elsarticle}

\usepackage{ycviu}
\usepackage{framed,multirow}

\usepackage{amssymb}
\usepackage{latexsym}

\usepackage{url}
\usepackage{xcolor}
\definecolor{newcolor}{rgb}{.8,.349,.1}

\usepackage{times}
\usepackage{epsfig}
\usepackage{graphicx}
\usepackage{amsmath}
\usepackage{amssymb}

\usepackage{booktabs}
\usepackage{paralist}
\usepackage{algorithmic,algorithm}

\usepackage[export]{adjustbox}
\usepackage{subcaption}

\journal{Computer Vision and Image Understanding}

\begin{document}

\setcounter{page}{1}

\begin{frontmatter}

\title{Performance Analysis and Robustification of Single-query 6-DoF Camera Pose Estimation}

\author[1]{Junsheng \snm{Fu}\corref{cor1}} 
\cortext[cor1]{Corresponding author: 
  Tel.: +46-709-637-400;}
\ead{junsheng.fu@tut.fi}
\author[1,2]{Said \snm{Pertuz}}
\author[3]{Jiri \snm{Matas}}
\author[1]{Joni-Kristian \snm{K\"am\"ar\"ainen}}

\address[1]{Tampere University of Technology, Department of Signal Processing
P.O. Box 553, FI-33101 Tampere, Finland}
\address[2]{Universidad Industrial de Santander, 680003 Bucaramanga, Colombia}
\address[3]{Czech Technical University in Prague, Faculty of Electrical Engineering, Technicka 2, 16627 Praha 6, Czech Republic}

\received{}
\finalform{}
\accepted{}
\availableonline{}
\communicated{}

\begin{abstract}
We consider a single-query 6-DoF camera pose estimation with reference images 
and a point cloud, 
i.e. the problem of estimating the position and orientation of a camera 
by using reference images and a point cloud.

In this work, we perform a systematic
comparison of three state-of-the-art strategies for 6-DoF camera pose
estimation, i.e. feature-based, photometric-based and mutual-information-based
approaches. The performance of the studied methods is evaluated on two 
standard datasets in terms of success rate, translation error and max 
orientation error. Building on the results analysis, we propose a hybrid approach 
that combines feature-based and mutual-information-based pose estimation methods 
since it provides complementary properties for pose estimation.

Experiments show that (1) in cases with large environmental variance, 
the hybrid approach outperforms feature-based and mutual-information-based 
approaches by an average of 25.1\% and 5.8\% in terms of success rate, respectively; 
(2) in cases where query and reference images are captured at similar 
imaging conditions, the hybrid approach performs similarly as the feature-based 
approach, but outperforms both photometric-based and mutual-information-based 
approaches with a clear margin; (3) the feature-based approach is consistently 
more accurate than mutual-information-based and photometric-based approaches when
at least 4 consistent matching points are found between the query and 
reference images.

\end{abstract}

\begin{keyword}
\KWD camera pose estimation\sep 3D point cloud\sep Hybrid method\sep Image feature\sep Photometric matching\sep Mutual information
\end{keyword}

\end{frontmatter}
%\articleinfobox

%\linenumbers

\section{Introduction}
\label{sec1}

Camera pose estimation is a fundamental technology for various
applications, such as augmented reality~\citep{hololens}, virtual
reality~\citep{ohta2014mixed}, and robotic
localization~\citep{castellanos2012mobile}. The aim of 6 degrees of
freedom (DoF) camera pose estimation is to find the 3-DoF location and
3-DoF orientation of the query image in a given reference coordinate
system. In the literature, the classical approach for 6-DoF camera
pose estimation is to register a 2D query image with previously
acquired reference data, which often consist of a set of reference
images and corresponding 3D point clouds. In practice, this is a
fundamental yet challenging problem due to large displacements between
the query and reference images, as well as image variations caused by
changes in the appearance of the scenes, weather and lighting
conditions~\citep{Oxford_data, mods}. Depending on the way to compute the 6-DoF
camera pose for the query image, the state-of-the-art methods can be
divided into 2 main categories: \textit{direct} and \textit{indirect}
approaches. In our context, \textit{direct} approach means the 6-DoF camera 
pose is directly optimized by a cost function at the space of 
6D camera pose. For example, 6-DoF camera pose can be computed 
by directly minimizing a cost function which compares the query image 
with a rendered synthetic view from a 3D point cloud, and the rendered 
view can be determined by either gradient or grid search
~\citep{nid-slam,tykkala2013photorealistic,rgbd-slam,dtam2011}.
In the \textit{indirect} approach, the query image is registered to
the 3D point cloud by matching against the reference
images~\citep{mods, 2d_2d_2, 2d_2d_4, 2d_2d_3d}, and the reference images 
and the 3D point cloud are defined in the same world coordinate system. This 
\textit{indirect} approach can be considered as a combinatorial 
optimization method, because we need to find the 2D-3D correspondences 
between the query image and the 
3D point cloud for computing the 6-DoF camera pose.
Both \textit{direct} and \textit{indirect} approaches have shown good performance in different 
literatures and different datasets with different setting
~\citep{nid-slam, mods, 2d_2d_2}, but the relative performance of the \textit{direct} 
and \textit{indirect} approaches have not been intensively analyzed in the same working 
conditions with large real-life dataset.

%
%some authors argue that the direct method, has the 
%limitation of relying on well initialized reference images~\citep{nid-slam}. 
%However, in the literature 

% and 3D point cloud~\citep{2D_3D_1}, then a Perspective-n-Point solver~\citep{pnp_1, pnp_2} is applied to compute the relative 6-DoF camera pose between the query image and the reference 3D point cloud. 
%
%In the direct approach, the camera pose is directly optimized by minimizing a cost function when compare the query image with the synthetic view generated of the reference 3D point cloud~\citep{nid-slam, tykkala2013photorealistic, rgbd-slam}. Recent method~\citep{nid-slam} that directly minimizes the mutual-information-based cost function instead of the photometric-based cost function, providing increased robustness to large appearance variation. However, this approach relies on well initialized reference images. 

Even though both the \textit{indirect} and \textit{direct} approaches have been widely
utilized for 6-DoF pose estimation, we have identified two important
questions that warrant further research: first, there is still no
consensus in the community about which strategies yield the best
performance in real-life conditions where the appearance of the
reference and query images change significantly according to different
weather, lighting and season conditions. Second, in the literature,
pose estimation strategies are often assessed as a part of full
pipelines that involve additional pre- or post-processing steps,
\textit{e.g.} the incorporation of information from previous poses in
sequential data or global optimization strategies in simultaneous
localization and mapping approaches. As a result, the contribution of
pose estimation methods on the overall performance of the system, as
well as their response to different imaging factors, remains unclear. 
In order to tackle the aforementioned problems, we implemented and
studied three start-of-the-art camera pose estimation approaches, 
to estimate 6-DoF camera pose of a single-query image using reference images 
and 3D point clouds. Specifically, these 3 approaches consist of 1 \textit{indirect} 
approach: a feature-based approach~\citep{2d_2d_3d}, and 2 \textit{direct} approaches: 
a photometric-based method~\citep{tykkala2013photorealistic} 
and a mutual-information-based method~\citep{nid-slam}. 
The motivation of studying the 3 chosen approaches is that they are 
state-of-the-art, have good speed performance and are 
convenient to be implemented~\citep{nid-slam,tykkala2013photorealistic, 2d_2d_3d}. 
We perform a systematic and extensive experimental comparison of the studied 
approaches and analyze their performances.

Based on the obtained results, we propose a hybrid
approach, consisting of the fusion of the feature-based and mutual
information-based camera pose estimation methods, and present an
architecture for computing the 6-DoF camera pose from rough 2-DoF
spatial position estimates. Our \textbf{main contributions} can be
summarized as follows: 

\begin{itemize}
\item We perform an extensive comparison and analysis of three
  strategies for 6-DoF camera pose estimation: feature-based approach,
  photometric-based approach, and mutual-information-based approach. We
  find that the feature-based approach is more accurate than the 
  photometric-based and mutual-information-based approach 
  with as few as 4 consistent feature points between the query
  and reference images. However, we also found that the 
  mutual-information-based approach is often more robust and can provide a pose 
  estimate when the feature-based approach fails.

\item We propose a hybrid approach that combines feature-based and
  mutual-information-based approaches based on the number of the
  feature matches between the query and reference images. We
  experimentally demonstrate that the hybrid approach outperforms both
  the feature-based only or the mutual-information-based only
  approaches.  
  
\item All code of the 3 implemented camera pose estimation methods and 
the performance evaluations will be made public.   

\end{itemize}

We evaluate the performance of the hybrid approach by implementing an
architecture that allows computing camera pose with multiple reference
images and allows to naturally integrate and refine pose priors in
large uncertainty cases. For the experiments, we used two
publicly available datasets: the KITTI dataset~\citep{kitti} and Oxford RobotCar
Dataset~\citep{Oxford_data}. The KITTI dataset provides 11 individual
sequences with ground truth trajectories. The recently released Oxford
RobotCar Dataset~\citep{Oxford_data}, contains many repetitions of a
consistent route and provides different combinations of weather,
traffic and pedestrians, along with longer term changes such as
construction and roadworks, which allows a more challenging evaluation
in extreme changing conditions. Our comparison shows how the hybrid
approach outperforms feature-based-only, photometric-based-only or
mutual-information-based-only approaches. Furthermore, the experiments
show the using multiple reference images improves the robustness of all pose
estimation pipelines.

\subsection{Related work}

Camera pose estimation using vision has received significant attention
in recent decades. We are focused on the case of registering a single
query image with one or several reference images and 3D point
clouds. The approaches can be divided into 2 main categories: the
\textit{indirect} approach~\citep{2d_2d_4, 2d_2d_3d} and the \textit{direct}
approach~\citep{nid-slam, tykkala2013photorealistic, rgbd-slam}.   

The \textit{indirect} approaches establish 2D-3D correspondences between the query
image and the 3D point cloud. The reference images and the 3D
point cloud are pre-registered, so the 2D-3D correspondences are achieved
by establishing 2D-2D correspondences between the query
image and the reference images. Specifically, the query image is
registered with the reference images by utilizing feature detectors
for finding the useful image structures for localization,
e.g. corners~\citep{rosten2006corner, mikolajczyk2004scale},
blobs~\citep{lowe1999sift,bay2006surf, kadir2001saliency} or
regions~\citep{matas2004mser, tuytelaars2000wide,
  tuytelaars2004matching, mori2004recovering}. Then feature
descriptors~\citep{calonder2010brief, rublee2011orb,
  leutenegger2011brisk, alahi2012freak, lowe1999sift, bay2006surf,
  dalal2005HOG, tola2010daisy, ambai2011card} are used to provide
robust representation regardless of appearance changes due to
different viewpoints, weather, lighting, etc. Given the set of 2D-3D
correspondences, a Perspective-n-Point solver~\citep{pnp_1, pnp_2} and 
RANSAC~\citep{ransac, pnp_1} are
applied to compute the relative 6-DoF camera pose between the query
image and the reference 3D point cloud. Because different combinations of 2D-3D correspondences 
lead to different camera pose estimations, the \textit{indirect} approach
can be considered as a combinatorial optimization method.

The \textit{direct} approaches compute the 6-DoF camera pose by minimizing a cost
function directly at the space of 6D camera pose~\citep{nid-slam,
tykkala2013photorealistic, rgbd-slam,dtam2011}, and do not need to extract local features of images. One commonly used 
cost function is photometric error between 
the query image and the reference view, where the reference view can be 
generated from the reference 3D point cloud~\citep{tykkala2013photorealistic, rgbd-slam, dtam2011}. 
The \textit{direct} photometric-based methods are easy to implement and have
good speed performance, however they are not robust to real-world global 
illumination changes~\citep{dtam2011}. A recent work~\citep{nid-slam}  
utilizes a mutual-information-based cost function for \textit{direct} 6-DoF
camera pose estimation outperforming both the feature-based and
photometric-based approaches in  
two challenging datasets with large image
variations. This mutual-information-based approach is
targeting on the application of SLAM problem, and it
relies on well-initialized reference image~\citep{nid-slam}. However, it is still
unclear what the performance of the mutual-information-based approach
would be without accounting for the initialization problem, where a
single query image is to be registered with no prior on the pose. 
To the best of our knowledge,
there is lack of prior art comparing the stand alone performance of
\textit{direct} and \textit{indirect} camera pose estimation approaches in this scenario.

\subsection{Overview}
\label{sec:methods}
Based on our literature review, we selected and implemented three 
state-of-the-art 6-DoF pose estimation methods: 
(1) \textit{indirect} feature-based
method~\citep{2d_2d_3d}, (2) \textit{direct} photometric-based
method~\citep{tykkala2013photorealistic} and (3) \textit{direct} 
mutual-information-based method~\citep{nid-slam}. 
We choose these 3 approaches because they have good 
performance and are convenient to be implemented. 
The details of these methods are presented in
Section~\ref{sec:pose_estimation_pipelines}. In order to conduct a
rigorous and systematic analysis of their practical performance, the
studied methods were compared in three different scenarios: the
single-reference case, the multi-reference case and the large uncertainty
case. Each one of the experimental setups for these 3 cases is
described in
Section~\ref{sect:comparative_methodology}. For the large uncertainty case,
we also present an architecture that allows the incorporation of
external pose information, e.g. GPS data. Experimental results on real datasets are 
presented in Sections~\ref{sec:experiments}. Based on the experimental results, we 
propose to integrate both \textit{direct} and \textit{indirect} methods into a 
\textit{hybrid approach} for an improved performance. 
The final discussion and the conclusion of this work are presented in
Section~\ref{sect:discussion} and~\ref{sect:conclusion} respectively.

%%%%%%%%%%%%%%%%%%%%%%%%%%%%%%%%%%%%%%%%%%%%%%%%%%%%%%%%%%%%%%%%%%%%%%%%%%%%%%%
\section{Evaluated pose estimation methods}
\label{sec:pose_estimation_pipelines}

\begin{figure}
  \begin{center}
    \includegraphics[width=1\linewidth]{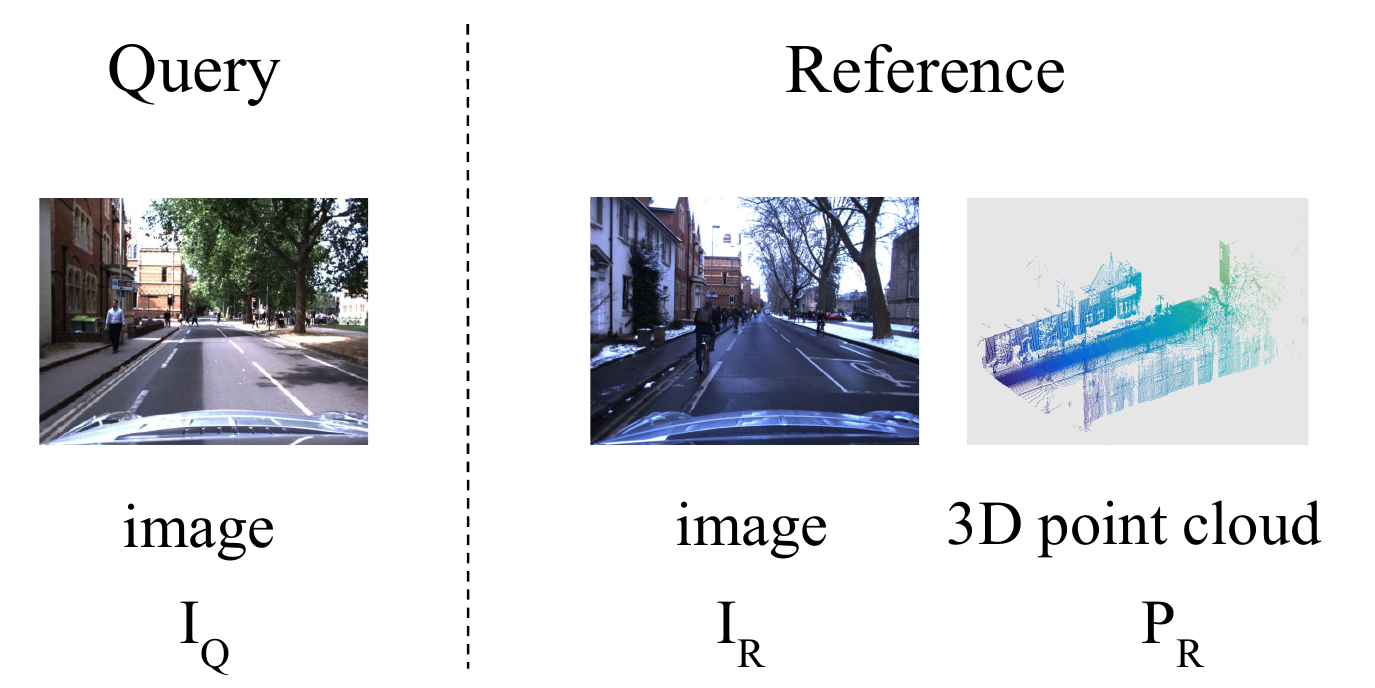}
  \end{center}
  \caption{Inputs for the pose estimation methods in the simplest scenario: 
    a query image $I_Q$ and a \textit{reference tuple} $(I_R, P_R)$,
    where $I_R$ is a single reference image and
    $P_R$ is the registered 3D point cloud associated to $I_R$. Both the the point cloud $P_R$
    and the camera pose of the reference image $I_R$ are defined in a common world coordinate system.}
  \label{fig:pipeline_single_case}
\end{figure}

The evaluated pose estimation methods in this work are: (1) \textit{indirect}
feature-basedmethod~\citep{2d_2d_3d}, (2) \textit{direct} photometric-based
method~\citep{tykkala2013photorealistic} and (3) \textit{direct} 
mutual-information-based method~\citep{nid-slam}. These three methods are good examples of 
\textit{direct} and \textit{indirect} approaches, presents have state-of-the-art  
performance and are convenient to be implemented. In this section, we describe each one of
the methods in the simplest scenario, where the inputs of 
all these three methodologies are a query image $I_Q$ and a single
\textit{reference tuple} $(I_R, P_R)$ that is formed by a reference
image $I_R$ and its registered 3D point cloud $P_R$, as illustrated
in Fig.~\ref{fig:pipeline_single_case}.

\subsection{Indirect feature-based (FB) pose estimation}
\label{sect:feature}

\begin{figure*}
\begin{center}
   \includegraphics[width=0.9\linewidth]{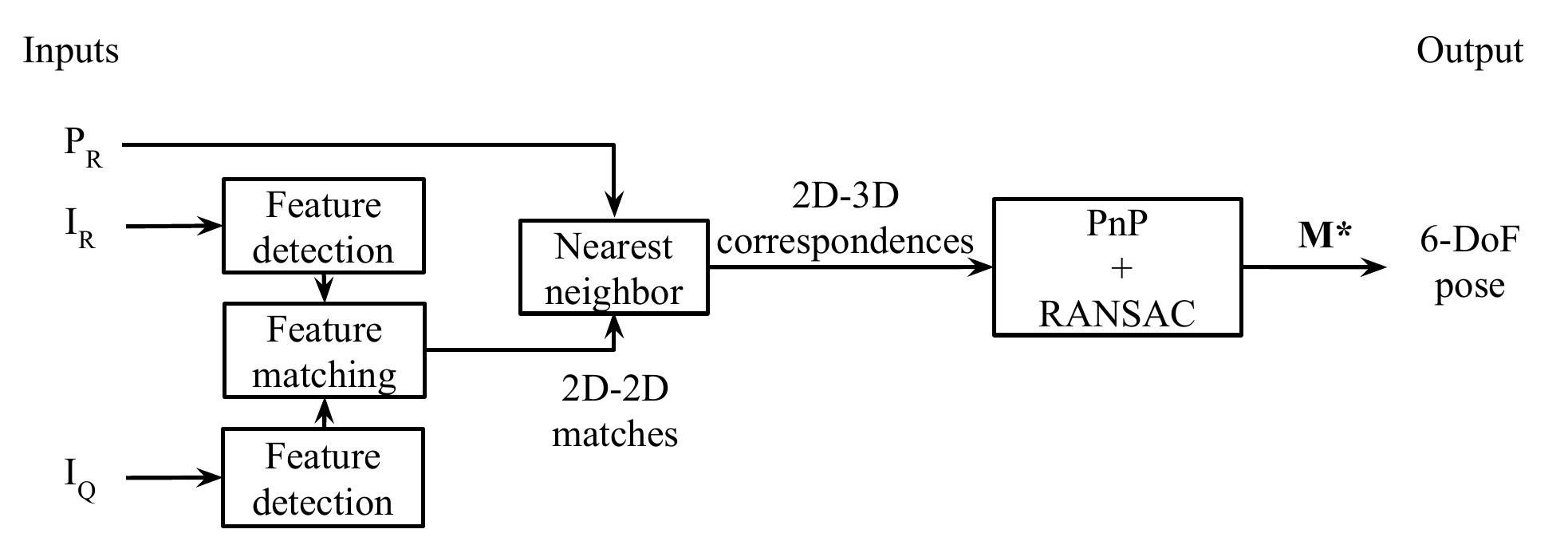}
\end{center}
   \caption{Block diagram of feature-based camera pose estimation. $I_Q$ 
     is the query image. The reference image $I_R$ and the 3D point 
     cloud $P_R$ are pre-registered and defined in the 
     world coordinate system. 
     $\textbf{M}^*$ is the estimated transformation matrix.
     For the detailed descriptions of each step see~\ref{append:fb}.} 
\label{fig:feature_pipeline}
\end{figure*}

A standard feature-based pose estimation method can be divided into four main
steps: $(1)$ feature detection, $(2)$ feature matching, $(3)$
2D-3D correspondences grouping, and $(4)$ Perspective-n-Point pose
estimation. The block diagram of this method is shown in
Fig.~\ref{fig:feature_pipeline}. In the first step, a feature detector
and a feature descriptor are applied to both query and reference
images to find interest-points or regions and form their descriptors
from pixels surrounding each detected region.
Secondly, based on the descriptors of the feature
points, 2D-2D correspondences are sought between query and reference
images with a feature matcher. Thirdly, since the 3D point cloud is
registered with the reference image, the 2D-3D correspondences between the query 
image and the 3D point cloud can be computed through the 2D-2D correspondences between 
the query and reference image. Finally, a Perspective-n-Point solver~\citep{pnp_2} and 
RANSAC~\citep{ransac,pnp_1} are applied for computing the 6-DoF camera
pose of the query image. 
The algorithm and implementation details 
of each stage of the feature-based pose estimation can be found
in~\ref{append:fb}.

\subsection{Direct photometric-based (PB) pose estimation}
\label{sect:photometric}

\begin{figure*}
\begin{center}
   \includegraphics[width=0.9\linewidth]{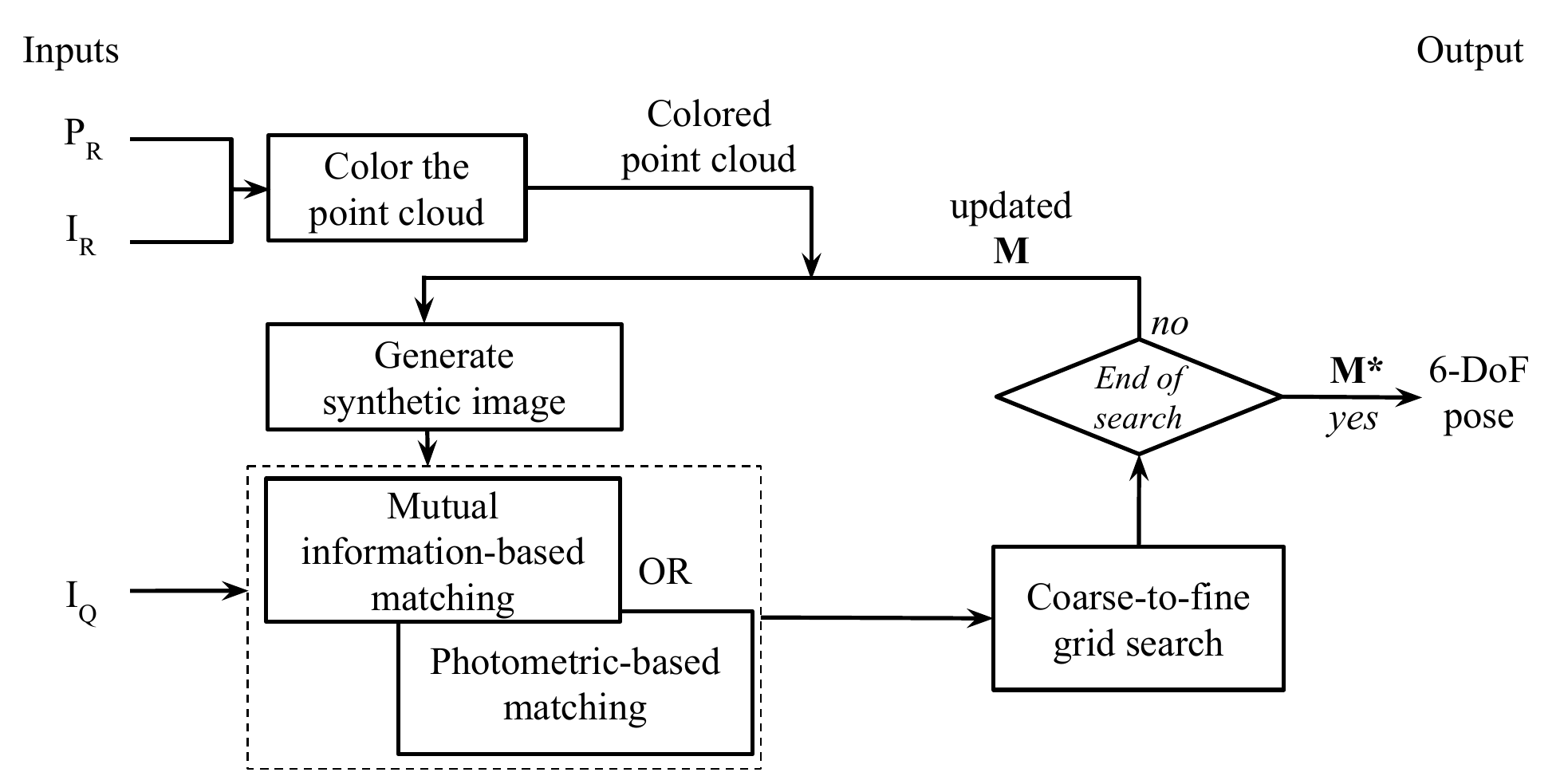}
\end{center}
   \caption{Block diagram of direct photometric-based and mutual 
     information based camera pose estimation. $I_Q$ 
     is the query image. The reference image $I_R$ and the 3D point 
     cloud $P_R$ are pre-registered and defined in the 
     world coordinate system.      
     $\textbf{M}^*$ is the estimated transformation matrix.
     For the detailed descriptions of each step see~\ref{append:pm}.} 
\label{fig:dense_pipeline}
\end{figure*}

\begin{figure*}
\begin{center}
   \includegraphics[width=0.9\linewidth]{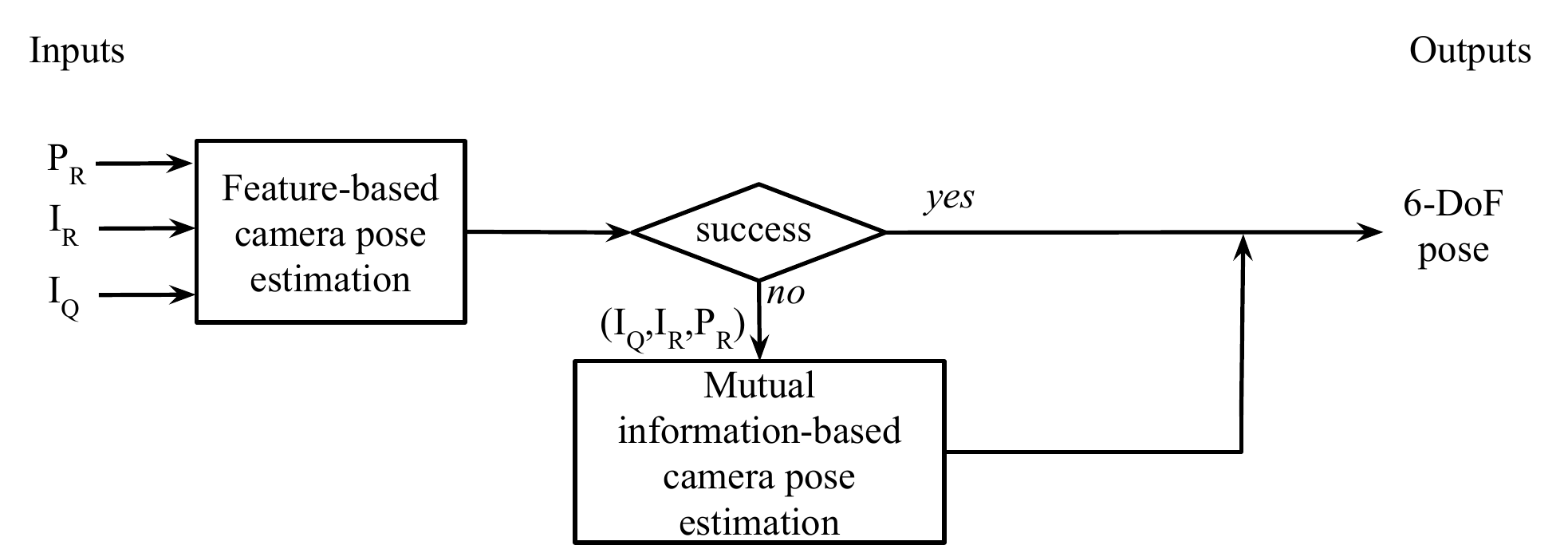}
\end{center}
   \caption{Block diagram of hybrid approach of camera pose
     estimation. $I_Q$ 
     is the query image. The reference image $I_R$ and the 3D point 
     cloud $P_R$ are pre-registered and defined in the 
     world coordinate system.} 
\label{fig:hybird_pipeline}
\end{figure*}

The \textit{direct} photometric-based approach~\citep{tykkala2013photorealistic} is defined as a direct minimization 
of the cost function at the space of 6D camera pose, and it does not need 
to extract local features. The pixel intensities of 
the query image and rendered synthetic view from the 3D point cloud are
directly compared in the cost function~\citep{tykkala2013photorealistic}. 
The photometric-based approach can be divided into three main
steps:  
$(1)$ synthetic image generation, $(2)$ photometric matching, and
$(3)$ coarse-to-fine search.

%\textit{\textcolor{red}{(Consider to use a algorithm table to describe it.)}}\\
The block diagram of this method is shown in Fig.~\ref{fig:dense_pipeline}. 
In summary the algorithm works as follows: firstly, for rendering a
{\em colored 3D point cloud} must be generated. This is generated
by projecting each 3D point of the cloud $P_R$ to
the reference image frame and then assigning colors from the points of
the reference image at that location.
Subsequently, we generate a
synthetic image by projecting the colored 3D point cloud into an image plane,
where the transformation matrix of the reference image is used as the
initial matrix. Then, we optimize the transformation matrix \textbf{M}
by a grid search. In the end, the 6-DoF camera pose
is obtained from the final transformation matrix $\textbf{M}^*$. It should
be noted that in common tracking applications where transformation baseline is small, fast
optimization can be implemented by using Jacobian and gradient-based
optimization~\citep{tykkala2013photorealistic}.
However, in the case of big appearances changes between the query and references images, 
the gradient tends to go to local minimum, so we conduct a grid search in our experiment.

A more detailed description of the
stages and implementation details of the photometric-based pose
estimation method can be found in~\ref{append:pm}.

% one possible alternatives, epnp

%
%
\subsection{Direct mutual-information-based (MI) pose estimation}
\label{sect:MI}

%\begin{figure}
%\begin{center}
%% \fbox{\rule{0pt}{2in} \rule{0.9\linewidth}{0pt}}
%   \includegraphics[width=1\linewidth]{source/pipeline_mi}
%\end{center}
%   \caption{Block diagram of mutual-information-based camera pose
%     estimation, where $I_Q$ is the query image, $I_R$ is the
%     reference image, and $P_R$ is the 3D point cloud.} 
%\label{fig:MI}
%\end{figure}
The \textit{direct} mutual-information-based approach~\citep{nid-slam} is a \textit{direct}
method similar as the photometric-based pose estimation, and it has more robust
similarity measurements. Because the above described \textit{direct} photometric-based 
pose approach is sensitive to photometric changes, e.g. due to illumination change.
To compensate these effects, a more robust
similarity measure -- {\em Mutual Information} (MI)~\citep{MI_1}  -- that can
be used to replace the direct pixel-based photometric-error in the
pose estimation cost function. The \textit{mutual information} is the measure of
the mutual dependence between two variables and can be used over
different modalities, and mutual-information-based image registration approaches 
are widely used in
medical image registration over different modalities~\citep{jbet2013121}. In turn, 
the normalized mutual information has the advantage that its values are in the
bounded range of $[0,1]$ ~\citep{MI_1}. 

The \textit{direct} mutual-information-based approach is similar to the \textit{direct} photometric-based
 pose approach in Section~\ref{sect:photometric}, with the main
difference being that in the cost function
the \textit{normalized mutual information} (NMI) is used instead of
the photometric error (see
Fig.~\ref{fig:dense_pipeline}).
Specifically, mutual
information based pose estimation is formulated as a minimization
problem as: 
\begin{equation}
\textbf{M}^{*} = \arg \min_\textbf{M} 1- \textit{NMI}(I_{Q},I_{S}),
\label{eq:mi_nid1}
\end{equation}
%\begin{equation}
%\textbf{M}^{*} = \arg \max_\textbf{M} NMI(I_{Q},I_{S}),
%\label{eq:mi_rmse1}
%\end{equation}
%
where $\textbf{M}^*$ is the estimated camera pose, $I_{Q}$ is the query image, 
$I_{S}$ is the synthetic image for which the generation process is described 
in~\ref{subsubsec:synthetic}, and the Normalized Mutual Information (NMI) is computed as: 

%\begin{equation}
%NID = 1 - NMI(I_{s};I_{q}) \enspace ,
%\label{eq:mi_1}
%\end{equation}

\begin{equation}
\textit{NMI}(I_{S},I_{Q}) = \dfrac{\textit{MI}(I_{S},I_{Q})}{max(H(I_{S}), H(I_{Q}))} \enspace
\label{eq:mi_2}
\end{equation}
with
\begin{equation}
MI(I_{S},I_{Q}) = H(I_{S}) + H(I_{Q}) - H(I_{S}, I_{Q}) \enspace ,
\label{eq:mi_3}
\end{equation}
where $H(I_{S}, I_{Q})$ is the joint entropy of $I_{S}$ and $I_{Q}$,
$H(I_{S})$ and $H(I_{Q})$ are the marginal entropies of $I_{S}$ and
$I_{Q}$, and $MI(I_{S},I_{Q})$ is the mutual information between
$I_{S}$ and $I_{Q}$.

\subsection{Hybrid (HY) pose estimation}
\label{sect:hybrid}

The hybrid approach for camera pose estimation takes the advantages of
both \textit{indirect} feature-based pose estimation and
\textit{direct} mutual-information-based pose estimation. This method
is inspired by the strong empirical evidence in our experiments that: (1) the 
feature-based method is superior in accuracy
if a sufficient number of matches can be found (see detail at Section
~\ref{subsec:exp_single_reference} and \ref{subsec:exp_large_uncertainty});  
(2) the feature-based approach can completely fail where mutual-information-based
approach can still provide a moderate
estimate. Therefore, our hybrid approach first executes the feature-based
method and if that fails ($< 4$ consistent 2D-3D correspondences)~\citep{pnp_1, pnp_2}, then
switches to the MI-based method.

Given one query image $I_Q$ and one \textit{reference tuple} $(I_R,
P_R)$ (see definition at Fig.~\ref{fig:pipeline_single_case} and Section
~\ref{sec:pose_estimation_pipelines}), 
a feature detector is firstly applied to both the query image $I_Q$ and reference image 
$I_R$ , and then we apply feature matching to get 2D-2D matched features.
Since the point cloud $P_R$ is registered with the reference image $I_Q$, the 2D-3D
correspondences can be found. Then a PnP solver~\citep{pnp_2} and RANSAC~\citep{pnp_1} are applied to the 2D-3D
correspondence. For the PnP solver~\citep{pnp_2}, at least 4 consistent 2D-3D correspondences pairs are required. 
If the camera pose of the query image cannot be estimated due to
less than 4 2D-3D correspondences~\citep{pnp_1, pnp_2}, 
the \textit{direct} mutual-information-based pose estimation is adopted to compute the
camera pose. The block diagram of the hybrid
approach is shown in Fig.~\ref{fig:hybird_pipeline}.

%%%%%%%%%%%%%%%%%%%%%%%%%%%%%%%%%%%%%%%%%%%%%%%%%%%%%%%%%%%%%%%%%%%%%%%%%%%%%%%
\section{Comparative methodology}
\label{sect:comparative_methodology}
In this work, we systematically compare camera pose estimation
approaches in three stages: firstly, we compare the performance of
different pose estimation methods for single query image in the simplest scenario by using
one \textit{reference tuple}, defined in Fig.\ref{fig:pipeline_single_case} (methodology 
in Section~\ref{subsct:single_ref} and experimental results in 
Section~\ref{subsec:exp_single_reference}). 
Secondly, we increase the number of reference images and evaluate the improvement in 
accuracy for the studied approaches (methodology in Section~\ref{subsec:multi-ref} 
and experimental results in Section~\ref{subsec:exp_multiple_reference}). Thirdly, 
we evaluate the different
approaches with large uncertainties, where the reference images and
their corresponding 3D point clouds can be far away from the query
image (methodology in Section~\ref{sec:pose_large_uncertainties} and experimental results 
in Section~\ref{subsec:exp_large_uncertainty}).

\subsection{Single-reference pose estimation}
\label{subsct:single_ref}

\begin{figure}
\begin{center}
   \includegraphics[width=1\linewidth]{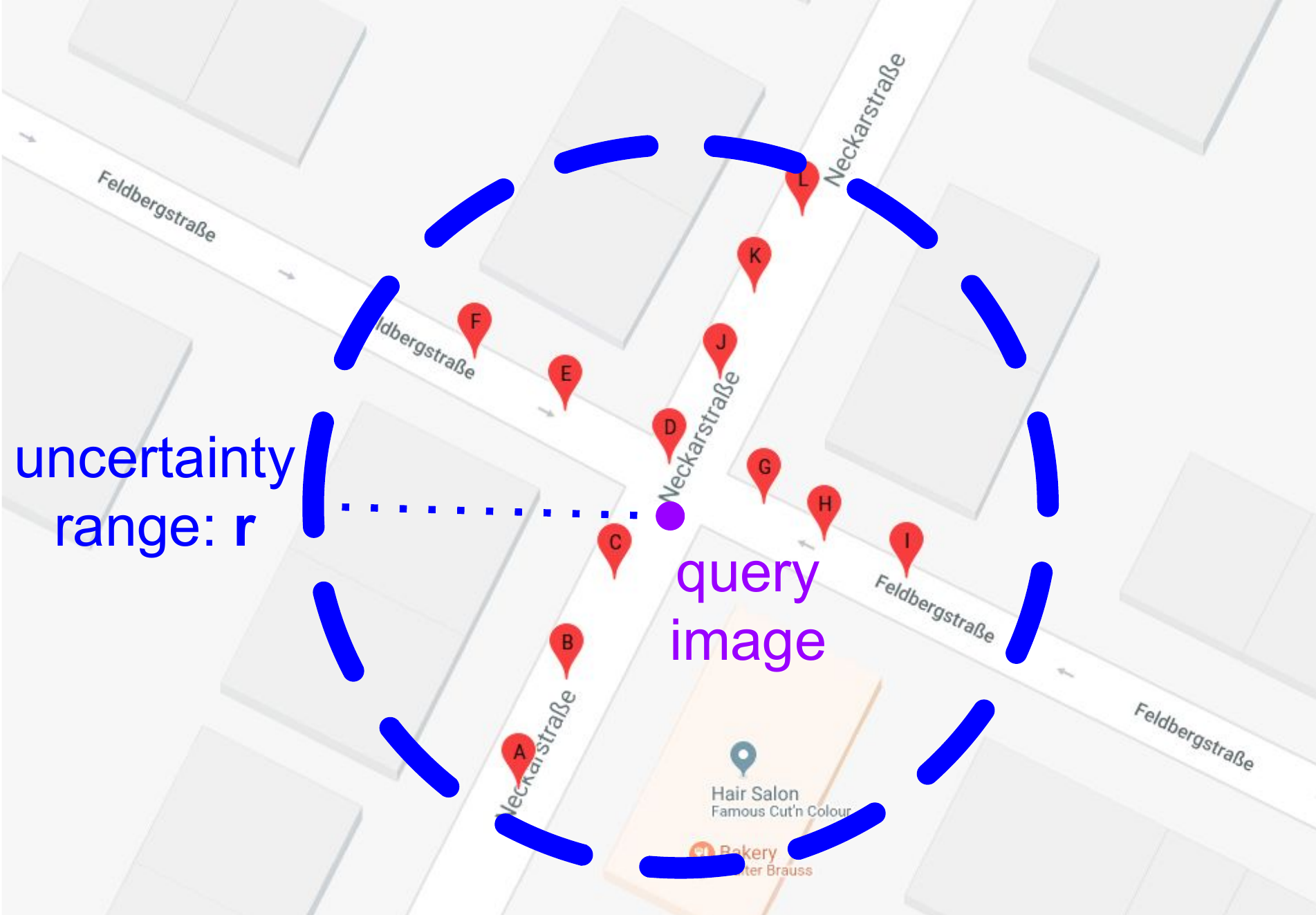}
\end{center}
\caption{This is the visualization of how a reference image is selected for camera pose estimation 
         in Section~\ref{subsct:single_ref}. The ground 
		 truth position of the query image is marked with a purple dot, and a circle around 
		 the purple dot represents the initial uncertainty of the query image's location. 
		 Within the initial uncertainty circle, one reference image is randomly selected among 
		 all possible candidates that are indicated with red markers from A to L.}
\label{fig:description_single}
\end{figure}

The aim of using a single reference image for different pose
estimation methods is to compare their performance at the most basic
level without pre- or post-processing steps. As illustrated in
Fig.~\ref{fig:description_single}, 
the experiment starts by firstly defining a uncertainty range with a radius of
$r$ around the ground truth location of the query
image, and $r$ represents the initial uncertainty of the query image's location. 
In this paper, initial uncertainty value $r$ is given by the author and it determines the 
region where the possible reference images can be chosen from.
The reference image is randomly selected in the region
within the circle, and the aim of introducing the random selection is
to evaluate how the studied algorithms respond to different
displacements between the query and reference images, because one concern
about the performance of pose estimation methods is the need of a
proper initialization (i.e., a good estimate of the current camera
location). Therefore, After randomly
selecting one reference image within the radius, the inputs
of the single-reference case are the query image $I_Q$ and
a \textit{reference tuple} $(I_R, P_R)$, where where $I_R$ is a single reference 
image and $P_R$ is its corresponding 3D point cloud (an example of 
\textit{reference tuple} is shown in Fig.\ref{fig:pipeline_single_case}). 
The quality of the estimated pose is then assessed in terms of the translation error and
rotation error (see Section~\ref{subsct:measures}).

\subsection{Multiple-reference pose estimation}
\label{subsec:multi-ref}

\begin{figure}
\begin{center}
   \includegraphics[width=1\linewidth]{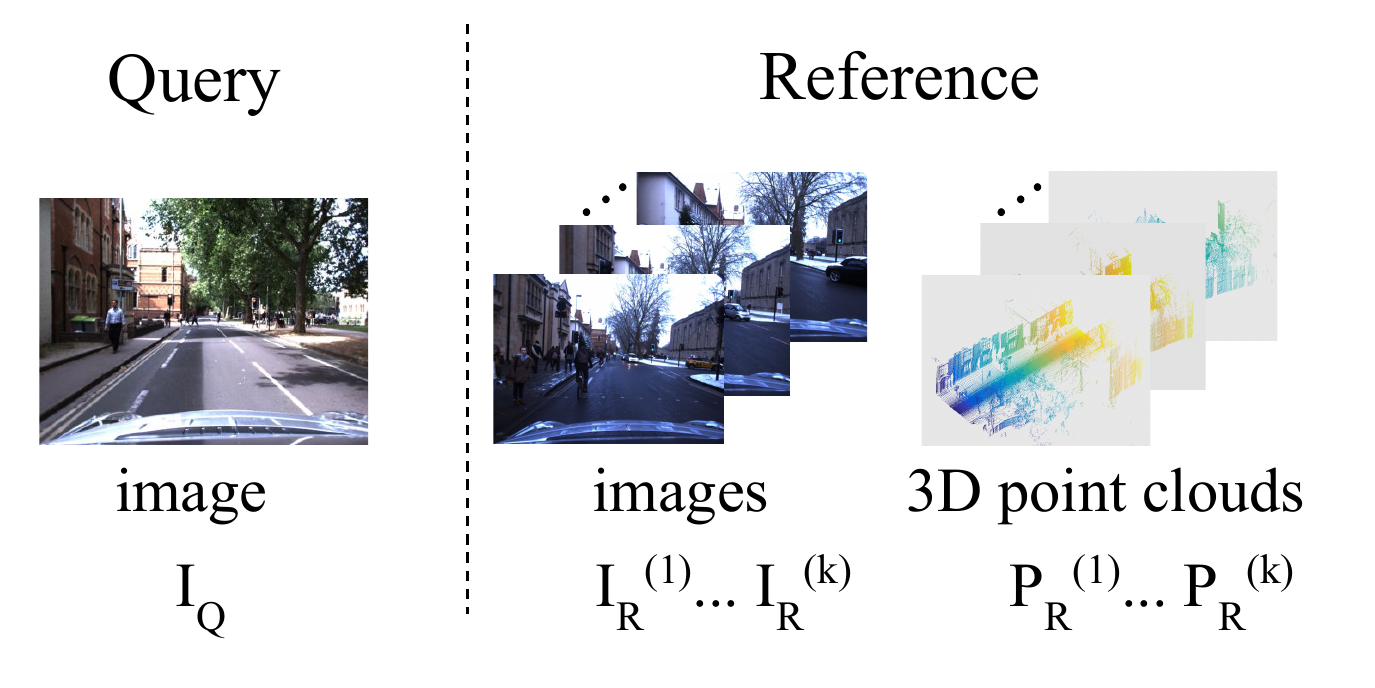}
\end{center}
   \caption{Example of inputs for multi-references case: one query
     image $I_Q$ and multiple \textit{reference tuples} $\{
     (I_{R}^{(1)},{P_{R}}^{(1)}), \ldots (I_{R}^{(k)}, {P_{R}}^{(k)})
     \}$ which consist of $k$ reference images and $k$ 3D point
     clouds.} 
\label{fig:multi_ref}
\end{figure}

In this section we explain the case of incorporating the information obtained
from multiple reference images to estimate the camera pose of a single query image. 
Therefore, the inputs are one query image and multiple \textit{reference tuples} 
which consist of $k$
pairs of reference images and their corresponding 3D point clouds, $\{
(I_{R}^{(1)},{P_{R}}^{(1)}), \ldots (I_{R}^{(k)}, {P_{R}}^{(k)}) \}$,
as shown in Fig.~\ref{fig:multi_ref}. The aim of using multiple reference images 
is to leverage the additional information of different reference images to improve
accuracy of the camera pose estimation. 

In the prior art, \citet{2d_2d_2} fuse multiple candidate camera poses
by: (1) averaging three rotation angles to compute the final rotation
matrix; (2) minimizing a geometry error term to estimate the final
translation. However, 3D point clouds are not utilized in their approach,
so from each of their candidate camera pose only a
line where the camera pose of the query image should lie on is obtained. In contrast,
in our approach, each reference image together with the 3D point
clouds are already sufficient to compute a unique 6-DoF camera pose
for the query image. Therefore, we have
considered 4 strategies, which can be easily adapted to different
camera pose estimation methods. 
\begin{enumerate}
\item Maximum number of matched features (\textit{maxf}): we match the query image
  with all the available reference images, and select the reference
  image with the most matched features after the feature matching
  stage (Section~\ref{subsec:feature_matching}). Then, we compute the
  camera pose of the query image with only the \textit{reference tuple}
  contains the selected reference image. The remaining processing steps are
  the same as in the camera pose estimation with single \textit{reference
    tuple} (see Section~\ref{sec:pose_estimation_pipelines}). 

\item Simple average (\textit{avg}): for each \textit{reference tuple} in $\{
  (I_{R}^{(1)},{P_{R}}^{(1)}), \ldots (I_{R}^{(k)}, {P_{R}}^{(k)}) \}$
  , we compute an individual candidate camera pose as described in
  Section~\ref{sec:pose_estimation_pipelines}. As a result, $k$
  candidate camera poses will be obtained. Each 6-DoF camera pose
  consists of a rotation matrix and a translation vector. We average
  the $k$ rotation matrices by firstly converting them to quaternions
  and then apply quaternion space
  interpolation~\citep{markley2007_Qaveraging}. As a result, the final
  rotation matrix is obtained from the interpolated quaternion, and
  the final translation vector can be computed by averaging all the
  translation vectors. 

\item Weighted average (\textit{wavg}): similarly as \textit{simple average}, this
  approach starts with $k$ individual candidate camera pose estimates
  obtained from each \textit{reference tuple}. Then we take a weighted
  average of these $k$ camera poses, and the weights are computed
  according to the number of the matched features between the query
  image and each reference image. 

\item Robust weighted average (\textit{r-wavg}): firstly we match the query image with
  all the available reference images and record the numbers of their
  matches. If the maximum number of matches is $K$ between the query
  and each reference image, we select those reference images with at
  least half of the maximum matches $K/2$. Then, we use them to compute the
  individual candidate camera poses and apply the weighted average
  over them. 

\end{enumerate}

\subsection{Camera pose estimation with large uncertainties}
\label{sec:pose_large_uncertainties}

\begin{figure}
\begin{center}
   \includegraphics[width=1\linewidth]{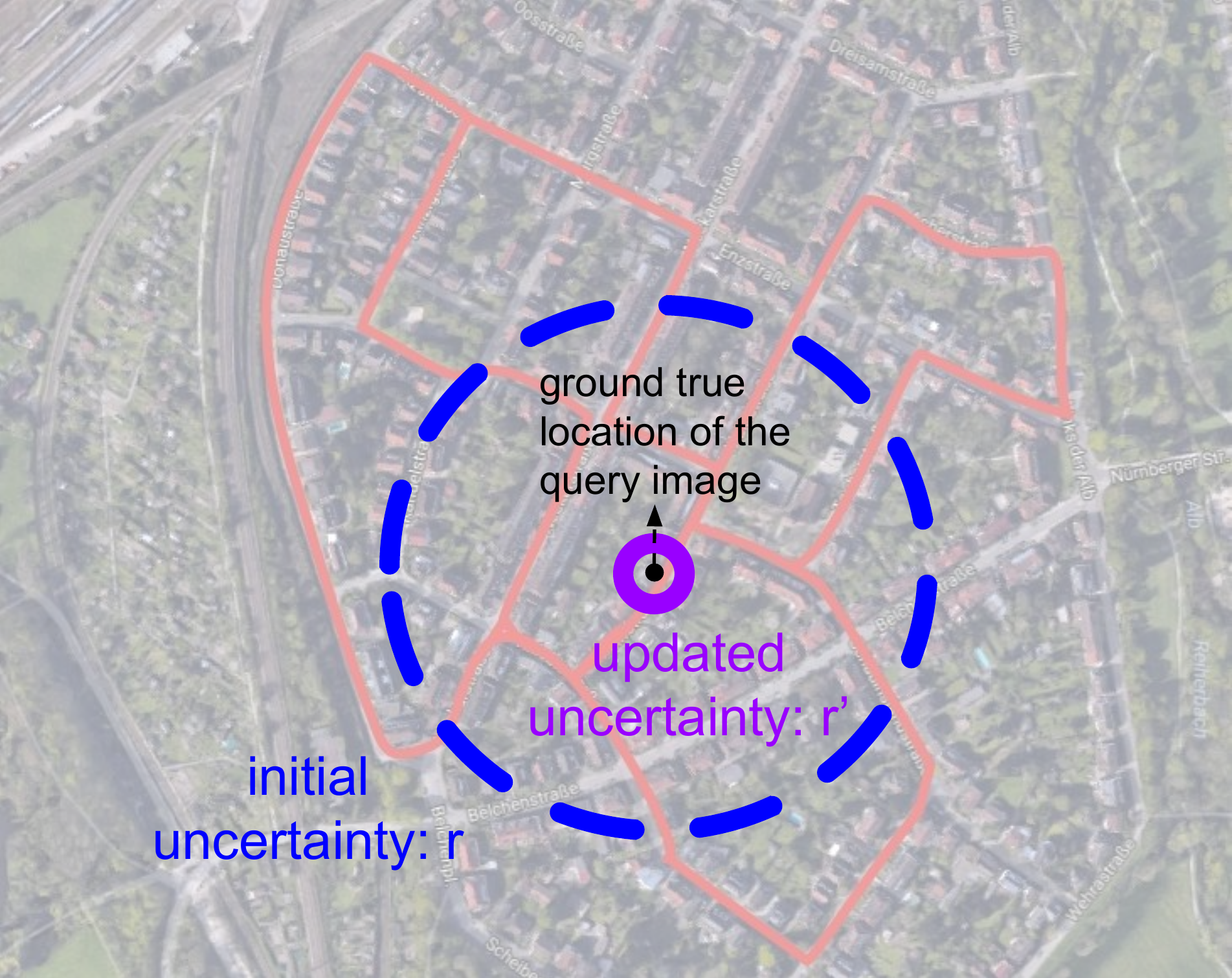}
\end{center}
\caption{ It is a visualization of applying an image retrieval method and a camera 
  pose estimation method to reduce the large position uncertainty of the query image. 
  The black dot represents the ground true location of the query image, the big blue dashed circle  
  shows the initial uncertainty and the small purple solid circle indicates the updated uncertainty
  after processing. The red route marked in the background is one of the KITTI dataset route.}
\label{fig:certainty}
\end{figure}

In real-life applications, the query image may or may not have a GPS tag, 
and even with a GPS tag, the precision of the GPS can be
poor~\citep{linegar2016made, gps1}. Therefore, the initial uncertainty radius $r$ of the query camera's location can
be as large as shown in Fig.~\ref{fig:certainty}. In the case of large uncertainty, choose the reference image by 
random selection is not practical anymore, but the use of image retrieval methods is a widely accepted practice. 
Therefore, we compare the performances 
of the studied pose estimation methods with large uncertainty, and evaluate how image 
retrieval improves their performance.

Specifically, for the case of the query
image with large initial uncertainty, image retrieval
methods such as \citep{2d_2d_2, philbin2007object, Radenovic-ECCV16,Iscen-CVPR17} 
are used to effectively identify a few good reference
images from the reference database. To replicate this procedure we
selected the method by~\citep{philbin2007object} which is easy to implement and 
performs the image retrieval task 
with large scale data by quantizing low-level image features based on randomized trees and 
using an efficient spatial verification stage to re-rank the results returned from our 
bag-of-words model. Furthermore, we apply all evaluated camera pose estimation methods with 
multiple reference images to compute the final 6-DoF camera pose.

%%%%%%%%%%%%%%%%%%%%%%%%%%%%%%%%%%%%%%%%%%%%%%%%%%%%%%%%%%%%%%%%%%%%%%%%%%%%%%%
\section{Experiments and results}
\label{sec:experiments}

%\section{Dataset}
%\label{sec:dataset}
\subsection{Datasets}

\begin{figure}
\begin{center}
   \includegraphics[width=1\linewidth]{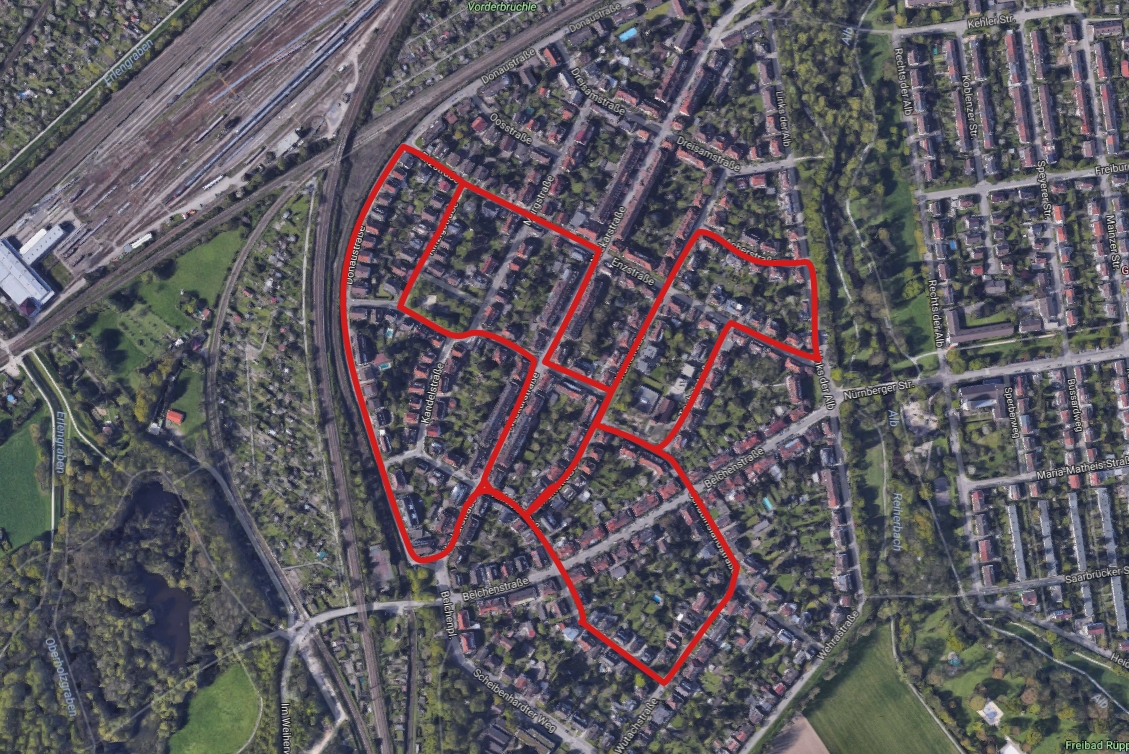}
\end{center}
\caption{The route of sequence 00 in Table~\ref{tab:kitti_summary} (KITTI dataset).}
\label{fig:kitti_example_route}
\end{figure}

In this work, experiments were conducted using two public datasets: the
KITTI Visual Odometry dataset~\citep{kitti} and the Oxford RobotCar
dataset~\citep{Oxford_data}. 

The KITTI dataset was captured by driving
around the mid-size city of Karlsruhe (Germany), in rural areas and on
highways. The accurate ground truth is provided by a Velodyne laser
scanner and a GPS localization system. There are 11 sequences in KITTI
Visual Odometry dataset with provided ground-truth
camera pose, and we use all of them in our experiments. All 11 sequences are summarized
in Table~\ref{tab:kitti_summary}. For each sequence, 3D point cloud $P_R$ is obtained 
from the provided LIDAR data, and both query image $I_Q$ and reference image $I_R$ are
from one monochrome camera (according to the author the monochrome camera is less noisy). 
One example of the sequence route for KITTI dataset is shown in Fig.~\ref{fig:kitti_example_route};

\begin{table}
\centering
\caption{Overview of the 11 sequences in the KITTI dataset~\citep{kitti}.}
\resizebox{0.5\textwidth}{!}{%
\begin{tabular}{ccccc}
\hline
\multirow{2}{*}{id} & \multirow{2}{*}{\# images} & \multirow{2}{*}{tag}  & \multirow{2}{*}{total length (km)} & \multirow{2}{*}{\begin{tabular}[c]{@{}l@{}}mean distance between \\ consequent images (m)\end{tabular}}\\ 
& & & & \\ \hline
00 & 4541 & urban  & 3.7 & 0.8\\
01 & 1101 & highway  & 2.5 & 2.2\\
02 & 4661 & urban  & 5.1 & 1.1\\
03 & 801 & urban  & 0.6 & 0.7\\
04 & 271 & urban  & 0.4 & 1.5\\
05 & 2761 & urban & 2.2 & 0.8 \\
06 & 1101 & urban  & 1.2 & 1.1\\
07 & 1101 & urban & 0.7 & 0.6 \\
08 & 4071 & urban  & 3.2 & 0.8\\
09 & 1591 & urban & 1.7 & 1.1 \\
10 & 1201 & urban  & 0.9 & 0.8\\ \hline
\end{tabular}
}

\label{tab:kitti_summary}
\end{table}

\begin{figure}
\begin{center}
   \includegraphics[width=1\linewidth]{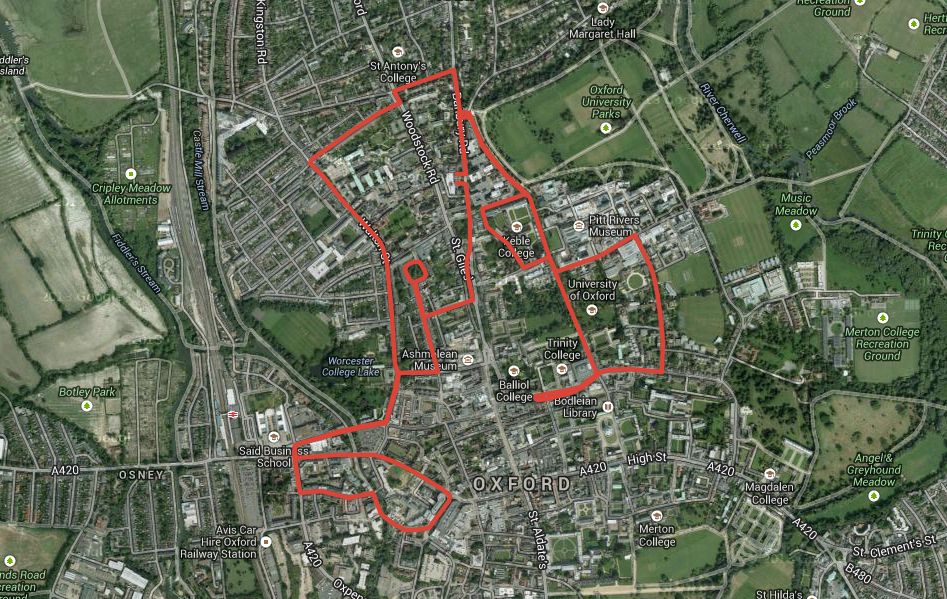}
\end{center}
\caption{The route used for all 5 sequences in Table~\ref{tab:oxford} (Oxford RobotCar dataset).}
\label{fig:oxford_example_route}
\end{figure}

\begin{figure*}
\begin{subfigure}{.19\textwidth}
  \centering
  \includegraphics[width=0.9\linewidth]{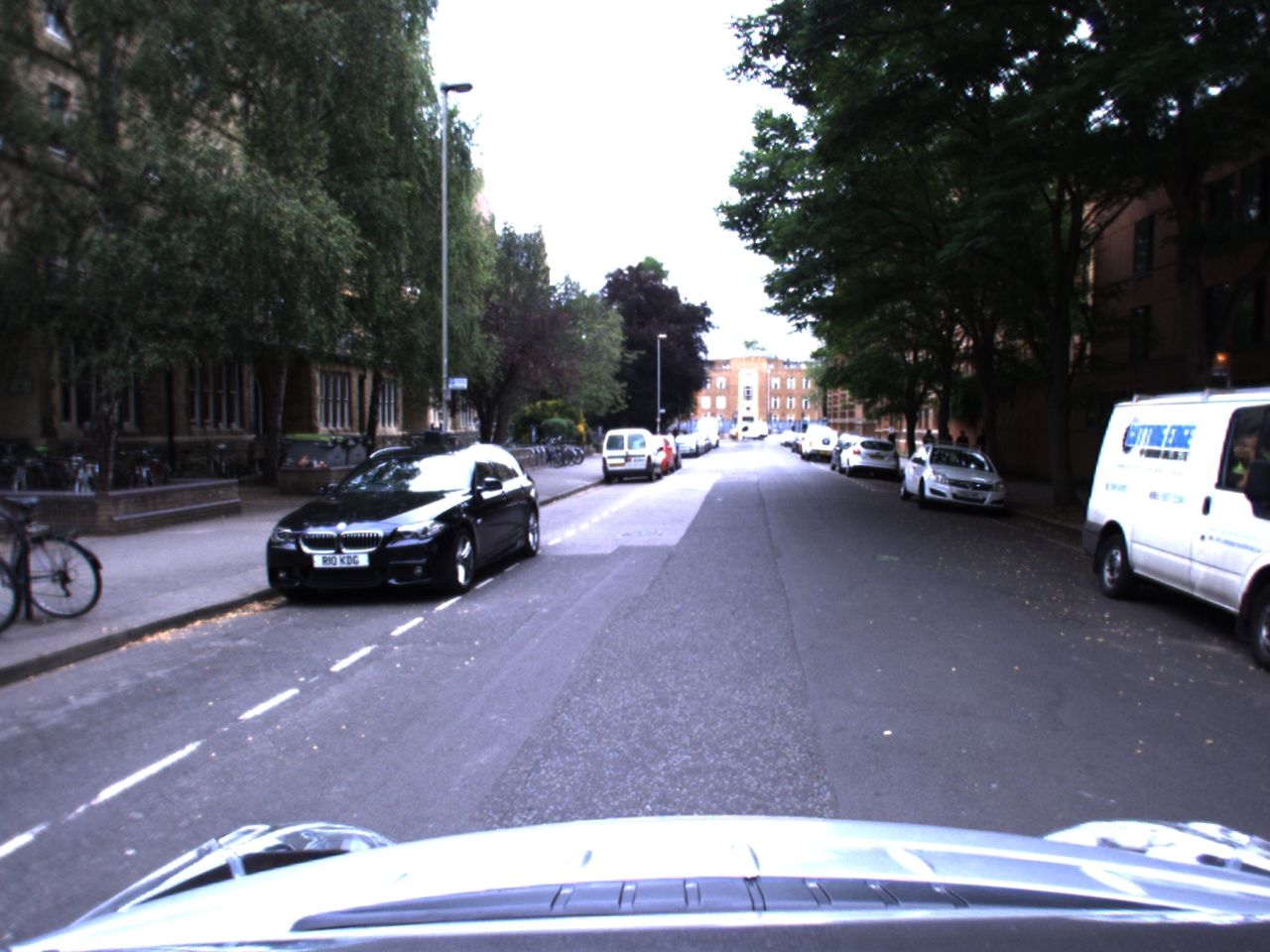}
  \caption{overcast}
  \label{fig:overcast}
\end{subfigure}%
\begin{subfigure}{.19\textwidth}
  \centering
  \includegraphics[width=0.9\linewidth]{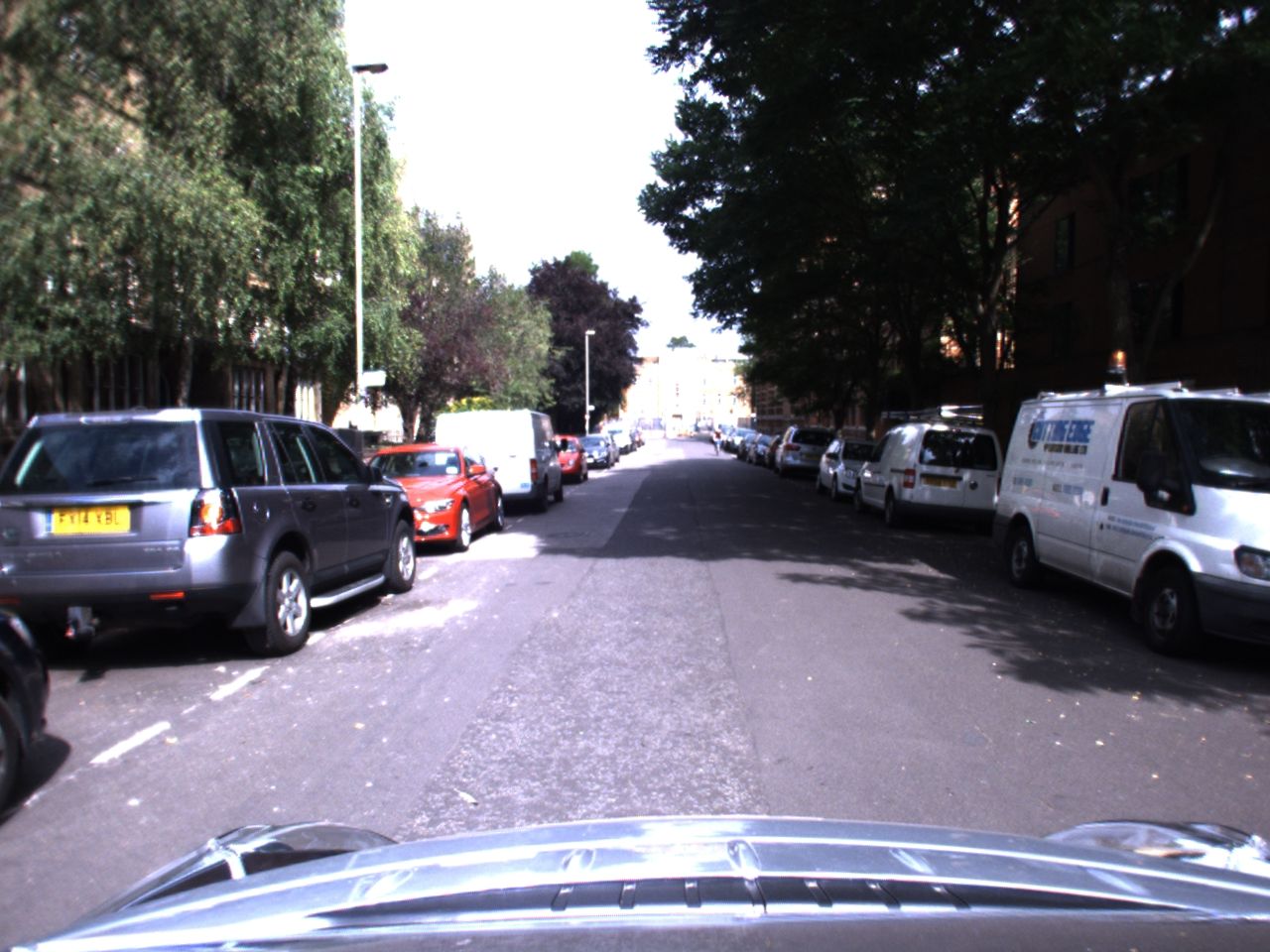}
  \caption{sun}
  \label{fig:sun}
\end{subfigure}
\begin{subfigure}{.19\textwidth}
  \centering
  \includegraphics[width=0.9\linewidth]{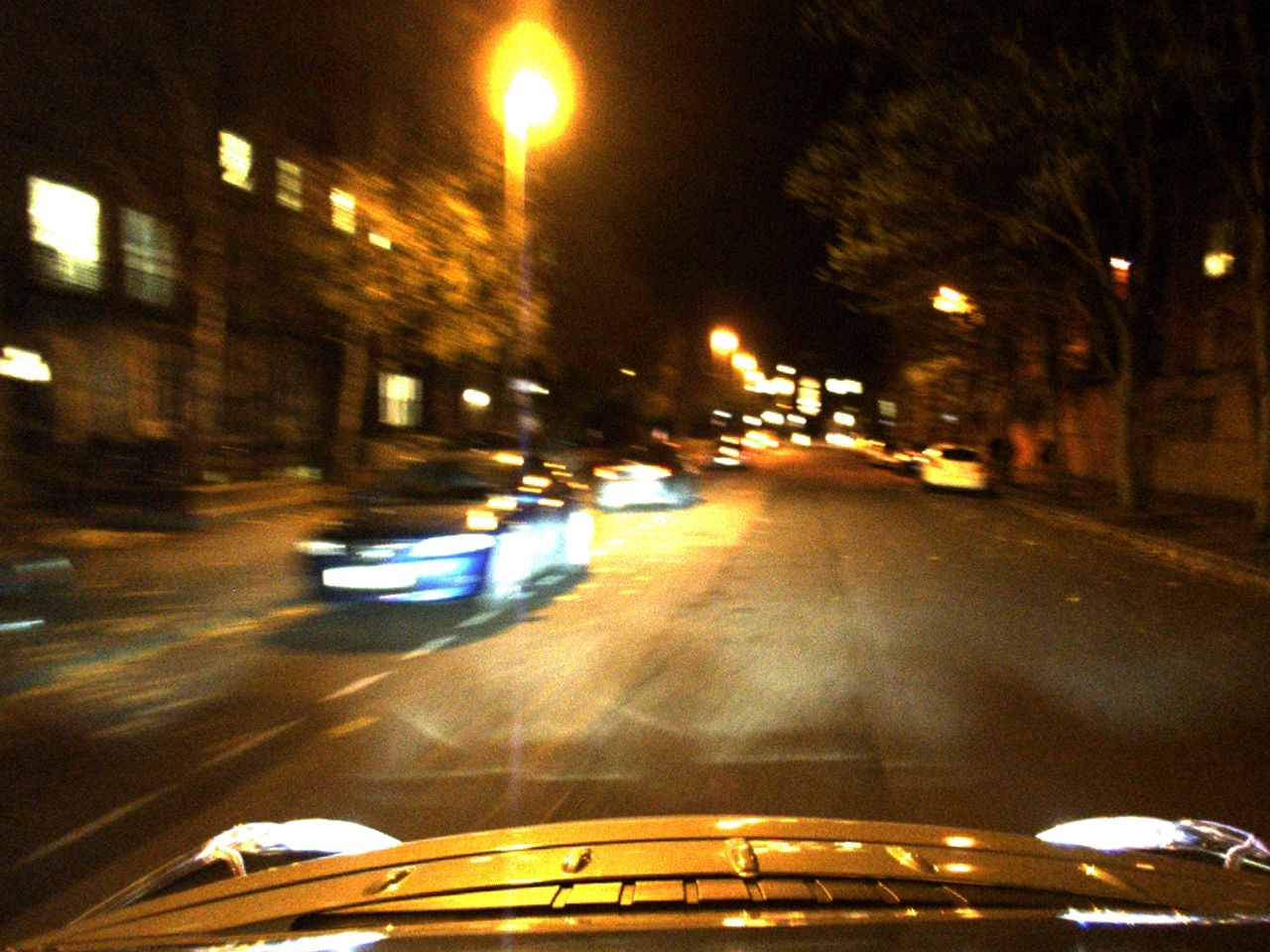}
  \caption{night}
  \label{fig:night}
\end{subfigure}
\begin{subfigure}{.19\textwidth}
  \centering
  \includegraphics[width=0.9\linewidth]{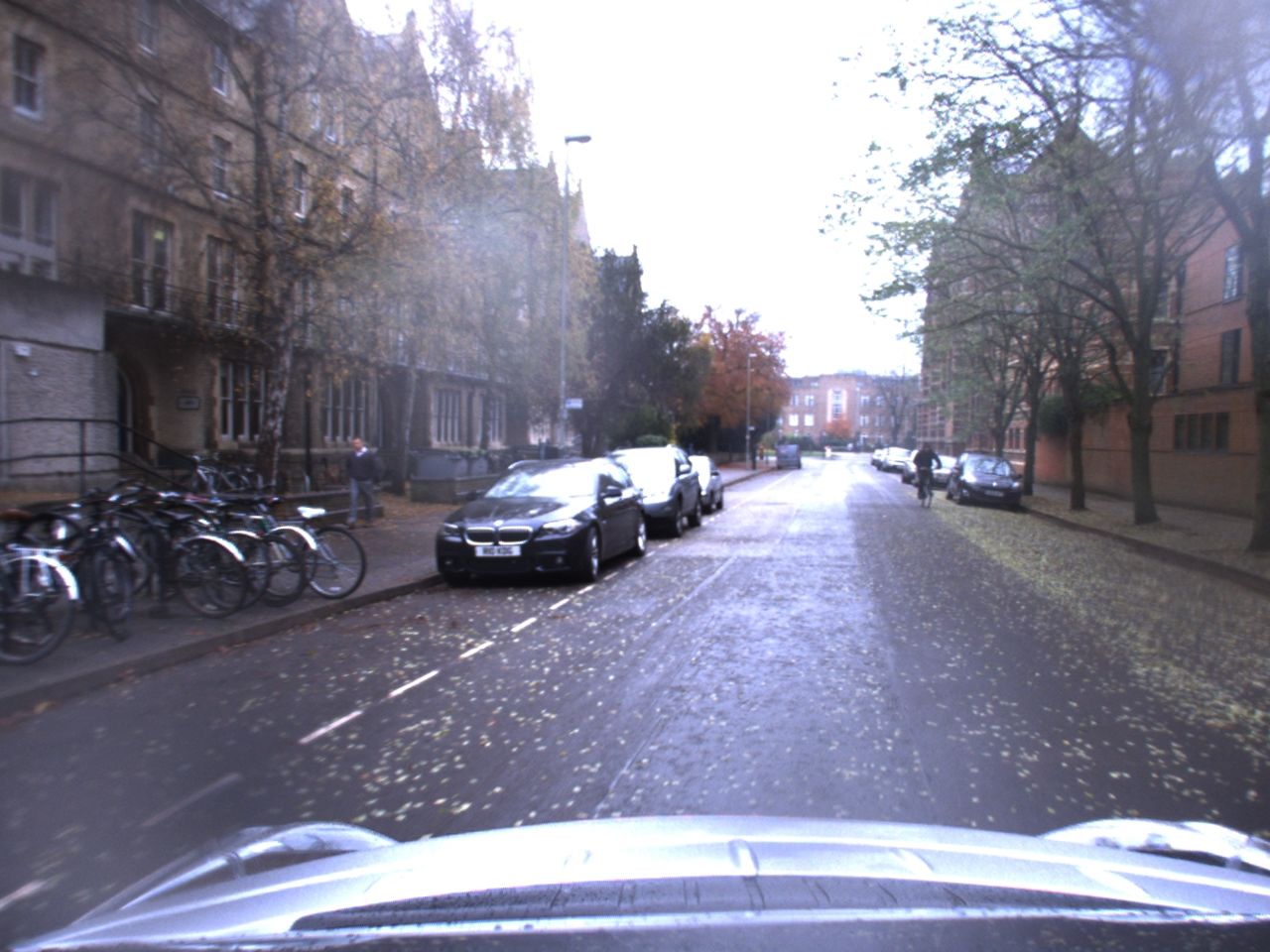}
  \caption{rain}
  \label{fig:rain}
\end{subfigure}
\begin{subfigure}{.19\textwidth}
  \centering
  \includegraphics[width=0.9\linewidth]{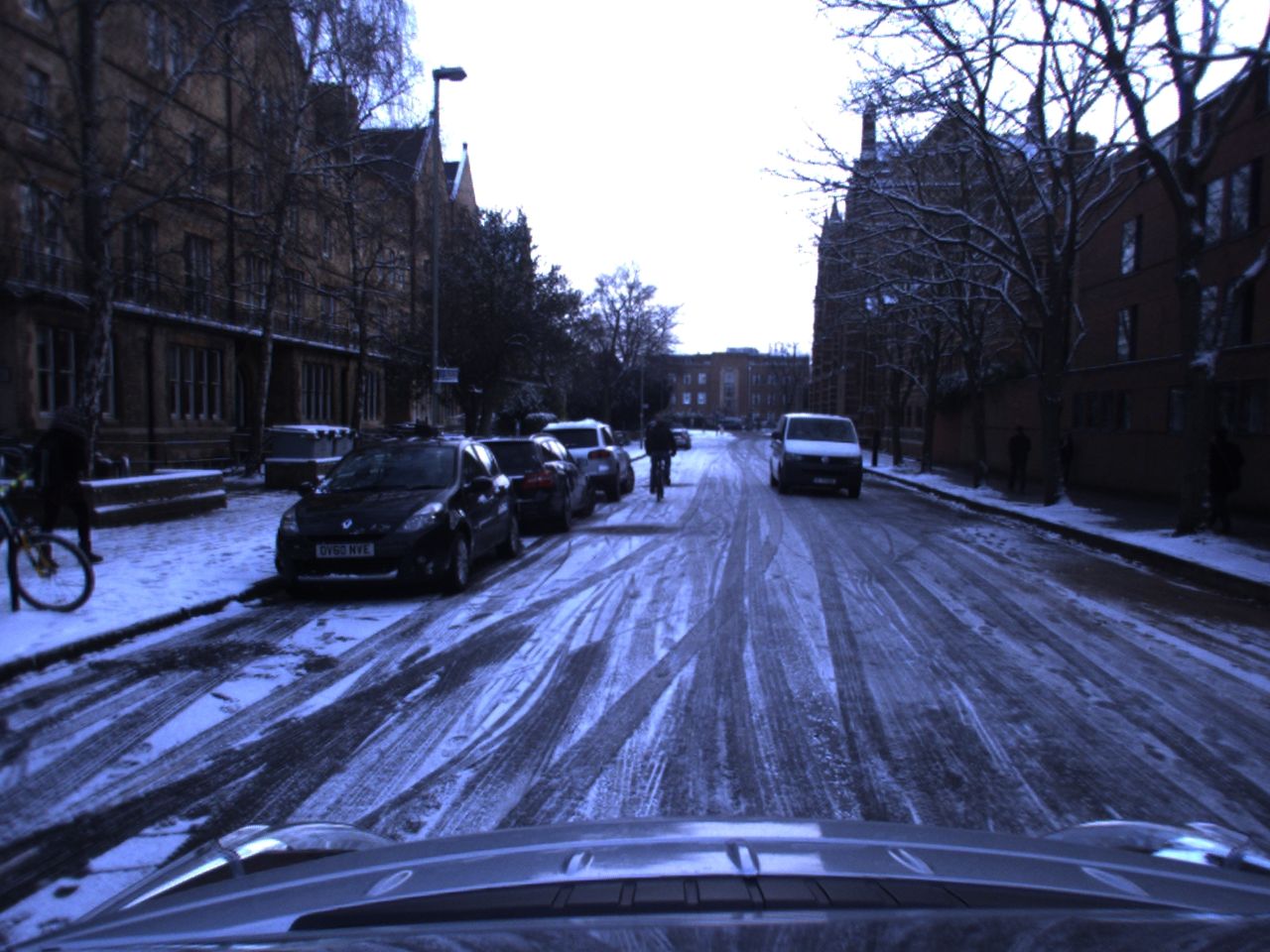}
  \caption{snow}
  \label{fig:snow}
\end{subfigure}
\caption{Appearance differences among the 5 sequences in Table~\ref{tab:oxford} (the images are roughly from the same location).}
\label{fig:oxford_examples}
\end{figure*}

The recently released Oxford RobotCar
dataset~\citep{Oxford_data} provides multiple traversals of the
same route and allows a more challenging evaluation in extreme
changing conditions, e.g. different time of the day, lighting and
weather condition. 5 sequences of the Oxford RobotCar dataset with
completely different environment conditions were selected for our
experiments. The sequence route is shown in
Fig.~\ref{fig:oxford_example_route} and example images from 5
sequences are shown in Fig.~\ref{fig:oxford_examples}. Similar to
KITTI dataset, 3D point cloud $P_R$ is from 3D LIDAR data, and query image and reference 
images are image taken from different traversals. The reported GPS information is treated 
as the ground-truth. For
efficiency, we reduced the number of images in each sequence by taking
1 image out of every 10 images and removed the beginning and ending
frames of each sequence where the car is usually parked resulting same views all the 
time. The resulting 5 sequences from
Oxford RobotCar dataset are summarized in Table~\ref{tab:oxford}. In
the case of Oxford dataset the query $I_Q$ and reference $I_R$ images
are taken from different traversals, and therefore give much more
realistic picture of pose estimation performance in real applications.

\begin{table}
\centering
\caption{Overview of 5 sequences with different environmental
  conditions in Oxford RobotCar dataset~\citep{Oxford_data}.} 
\resizebox{0.5\textwidth}{!}{%
\begin{tabular}{ccccc}
\hline
\multirow{2}{*}{id} & \multirow{2}{*}{\# images} & \multirow{2}{*}{tag}  & \multirow{2}{*}{total length (km)} & \multirow{2}{*}{\begin{tabular}[c]{@{}l@{}}mean distance between \\ consequent images (m)\end{tabular}}\\ 
& & & & \\ \hline
00 & 1916 & overcast  & 6.3 & 3.3\\
01 & 2873 & sun  & 8.6 & 3.0\\
02 & 2931 & night & 9.1 & 3.1 \\ 
03 & 2614 & rain  & 8.8 & 3.4 \\
04 & 3019 & snow  & 8.7 & 2.9\\ \hline
\end{tabular}
}

\label{tab:oxford}
\end{table}

%\section{Experiments}
%\label{sec:experiments}

%
%
\subsection{Performance measures}
\label{subsct:measures}

We use translation error, maximum orientation error and the success rate of 
each methods to compare the performance of the different approaches.  
\begin{enumerate}

\item The translation error is the absolute translation between the
  ground-truth location and the estimated location of the query image.  

\item  Based on the rotation matrix between the ground-truth camera
  pose and the estimated camera pose of the query image, we convert
  the rotation matrix into 3 Euler angles. Then the maximum absolute
  Euler angle is used as the maximum orientation error. 

\item Different camera pose estimation methods can fail to estimate a
  6-DoF camera pose of the query image under some circumstances,
  e.g. no enough feature matches between the query and reference
  images in \textit{indirect} feature-based approach, or searching 
  cannot converge within given search ranges in direct photometric-based
  approach. We classify the pose estimation failure as either self-reported by each
  method or the translation errors are greater than a predefined threshold
  (see details in the each experiment). The success rate of each methods 
  is the percentage of the successfully processed query images with 
  an valid camera pose as the output.    
\end{enumerate}

%For the experiments with the KITTI dataset, the query image and the reference images are from the same sequence, which allow us to test the camera pose estimation pipelines at an ideal condition, same time of the day, lighting and weather condition. 

\subsection{Experiments with single reference image}
\label{subsec:exp_single_reference}

We performed 12 experiments for KITTI dataset and 12 experiments for Oxford RobotCar
dataset. The goal of these experiments was to compare the performance
of different pose estimation methods under the single reference scenario.  
In Section~\ref{subsct:single_ref}, we discussed the camera pose
estimation for one query image $I_Q$ with single \textit{reference tuple}
$(I_R, P_R)$ where $I_R$ is a single reference image and $P_R$ is its
corresponding 3D point cloud. Similarly, we firstly gathered
all reference images within the uncertainty radius $r$ around the given 
query image's GPS location, and $r$ was varied between 10 to 25 meters, 
since most of the photos are taken in the streets of urban area, and 
these search ranges were used so that the reference image and 
query images would have some overlaps but without
being too close to each other. Within the gathered reference images,
we applied \textit{random selection} to choose one reference image $I_R$
and its corresponding 3D point cloud $P_R$. The reasons for using
\textit{random selection} is to evaluate how the studied algorithms 
respond to different displacements between the query and reference images
(i.e., different initializations).

The experiments with KITTI dataset tested the
performance of different camera pose estimation methods under
``ideal conditions'', \textit{i.e.} same time of the day, lighting
and weather condition. For the KITTI dataset listed in Table~\ref{tab:kitti_summary}, all 11
sequences have different routes, so each sequence was processed
individually. In other words, the query image and the reference images
come from the same drive. In order to separate the query and reference
images, we randomly selected 10\% of the images in one sequence as query
images, and the rest images from the same sequence were used as
reference images.

The experiments with Oxford RobotCar dataset tested the
performance of camera pose estimation methods at challenging
conditions since the query and reference data capture large variation
in appearance and structure of a dynamic city environment over long
periods of time. For the Oxford RobotCar dataset presented in Table~\ref{tab:oxford},
all 5 traversals with complete different environment settings 
share the same route. The sequences were processed
jointly in order to allow the query and reference images come from the
different sequences. For example, when the summer sunny sequence (01 in
Table.~\ref{tab:oxford}) was used for the reference images, the
winter snow sequence (04 in Table.~\ref{tab:oxford}) was used for the
queries.

\begin{figure*}
\begin{subfigure}{.5\textwidth}
  \centering
  \includegraphics[width=1\linewidth]{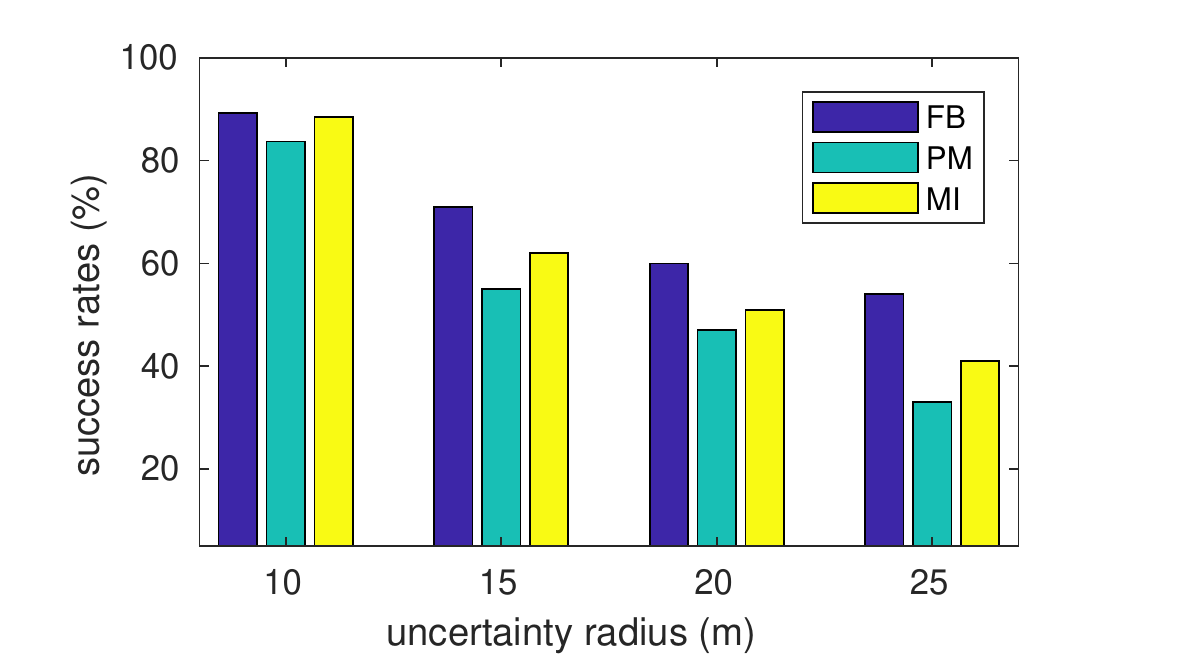}
  \caption{KITTI}
  \label{fig:kitti_exe1_rate}
\end{subfigure}%
\begin{subfigure}{.5\textwidth}
  \centering
  \includegraphics[width=1\linewidth]{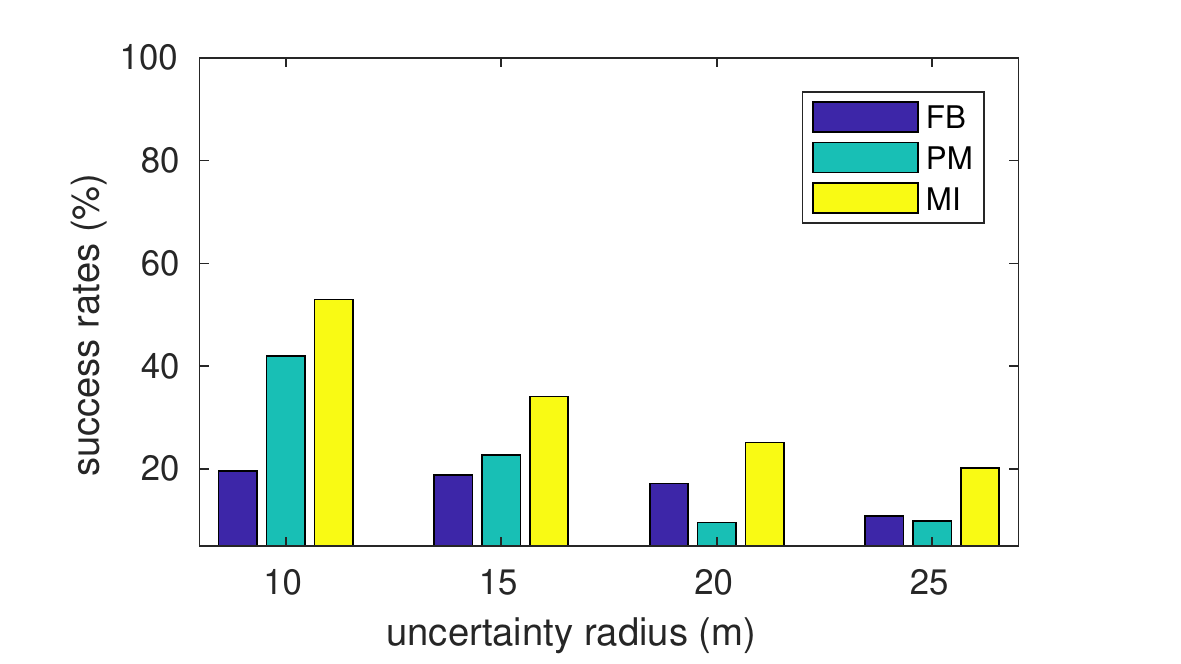}
  \caption{Oxford RobotCar}
  \label{fig:oxford_exe1_rate}
\end{subfigure}
\caption{\textbf{Success rate} comparison for three strategies with
  single reference image at different uncertainty ranges in two public
  datasets. (a): in the experiments with KITTI sequence 00, random
  10\% images in sequence 00 are used as query image and the rest are
  used as references. (b): in the experiments with two sequences in
  Oxford RobotCar sequences, summer sequence (01) is used as
  references and snow sequence (04) is used as query images.} 
\label{fig:exe1}
\end{figure*}

Table~\ref{tab:exe1} shows the translation errors and orientation errors 
of different pose estimation methods by using a single reference image. 
The results are reported in median values, and both the translation and 
orientation errors are
calculated based on the estimates obtained by the studied methods and
the ground-truth camera poses. The success rate of each pose
estimation method with Oxford RobotCar and KITTI datasets are shown
in Fig.~\ref{fig:exe1}. 

The main findings are that (1) the \textit{indirect} feature-based method (FB) 
is more accurate than \textit{direct} photometric-based (PM) and 
mutual-information-based (MI) approaches as long as there are at least 
4 consistent 2D-3D correspondences (4 is the minimum number of feature matches required 
to compute the camera pose by the PnP solver~\citep{pnp_2}); (2) but for the realistic Oxford 
RobotCar dataset, the success rate 
of feature-based method (FB) is clearly inferior to mutual-information-based method (MI).
Note that full analysis of the results is postponed
to the discussion section (Section~\ref{sect:discussion}).

\begin{table*}
\caption{\textbf{Translation error} (in meters) and \textbf{max orientation error} 
  (in degrees) comparison for three
  strategies using single reference image. For KITTI dataset, 454
  images (random 10\% of the whole sequence) in sequence 00 are use
  as queries, and the rest are used the reference images. For Oxford
  RobotCar dataset, summer sequence (01) is used as references and 302
  images (random 10\% of the winter sequence) from winter sequence are
  used as query images. Second row shows the number of images which
  are successfully processed by all three pipelines. Third row shows
  the percentage of the successfully processed images among all the
  testing images.}
\label{tab:exe1} 
  
\begin{subtable}{0.5\linewidth}\centering
{
\caption{KITTI sequence: translation error (m)}\label{tab:kitti_exe1_trans}
\resizebox{1\linewidth}{!}{ 
\begin{tabular}{lcccc}
\toprule
uncertainty radius (m) & 10 & 15 & 20 & 25 \\ 
\#images & 406  & 328  & 282  & 259  \\ 
 & (89\%) & (72\%) & (62\%) & (57\%) \\ \hline
FB~\citep{2d_2d_3d} & \textbf{0.13} & \textbf{0.40} & \textbf{0.48} & \textbf{0.30} \\
PM~\citep{tykkala2013photorealistic} & 1.44 & 6.66 & 7.77 & 14.85 \\
MI~\citep{nid-slam} & 1.56 & 5.41 & 6.15 & 10.26 \\ \bottomrule
%Random & 5 & 7.5 & 10 & 12.5 \\ 
\end{tabular}
}
}

\end{subtable}%
\begin{subtable}{0.5\linewidth}\centering
{
\caption{Oxford sequence: translation error (m)}\label{tab:oxford_exe1_trans}
\resizebox{1\linewidth}{!}{ 
\begin{tabular}{lcccc}
\toprule
uncertainty radius (m) & 10 & 15 & 20 & 25 \\ 
\#images & 67 & 60 & 53 & 38\\ 
 & (22\%) & (20\%) & (18\%) & (13\%)\\ \hline
FB~\citep{2d_2d_3d} & \textbf{2.77} & \textbf{2.48} & \textbf{2.40} & \textbf{2.91}\\
PM~\citep{tykkala2013photorealistic}  & 10.44 & 16.23 & 20.09 & 26.32\\
MI~\citep{nid-slam} & 8.71 & 13.36 & 16.27 & 14.94\\  \bottomrule
%Random & 1.56 & 3.16 & - & 4.32 \\ \bottomrule
\end{tabular}
}
}
\end{subtable} \\
\\
\begin{subtable}{0.5\linewidth}\centering
{
\caption{KITTI sequence: max orientation error (degree)}\label{tab:kitti_exe1_ori}
\resizebox{1\linewidth}{!}{ 
\begin{tabular}{lcccc}
\toprule
uncertainty radius (m) & 10 & 15 & 20 & 25 \\ 
\#images & 406  & 328  & 282  & 259  \\ 
 & (89\%) & (72\%) & (62\%) & (57\%) \\ \hline
FB~\citep{2d_2d_3d} & 1.76 & 3.83 & 5.42 & 3.33 \\
PM~\citep{tykkala2013photorealistic} & \textbf{1.07} & 2.40 & \textbf{3.37} & 3.12 \\
MI~\citep{nid-slam} & \textbf{1.07} & \textbf{2.30} & 3.45 & \textbf{2.70} \\ \bottomrule
%Random & 0.88 & 1.68 & 2.32 & 1.55 \\ \hline
\end{tabular}
}
}
\end{subtable}%
\begin{subtable}{0.5\linewidth}\centering
{
\caption{Oxford sequence: max orientation error (degree)}\label{tab:oxford_exe1_ori}
\resizebox{1\linewidth}{!}{ 
\begin{tabular}{lcccc}
\toprule
uncertainty radius (m) & 10 & 15 & 20 & 25 \\ 
\#images & 67 & 60 & 53 & 38\\ 
 & (22\%) & (20\%) & (18\%) & (13\%)\\ \hline
FB~\citep{2d_2d_3d} & \textbf{3.44} & \textbf{3.79} & 2.72 & 3.25\\
PM~\citep{tykkala2013photorealistic} & 3.48 & 5.82 & 2.64 & \textbf{1.88} \\
MI~\citep{nid-slam} & 6.16 & 4.00 & \textbf{2.42} & 1.93 \\  \bottomrule
%Random & 1.56 & 3.16 & - & 4.32 \\ \bottomrule
\end{tabular}
}
}
\end{subtable}
\end{table*}

\subsection{Experiments with multiple references images}
\label{subsec:exp_multiple_reference}

In this experiment, we evaluated the performance of different methods in the
multiple reference images setting for the both KITTI and Oxford RobotCar
dataset. The goal was to find efficient ways to incorporate the
information obtained from multiple reference images to improve the
camera pose estimation.

Similar to the single reference image case in
Section~\ref{subsec:exp_single_reference}, we first gathered the reference
images within a given uncertainty radius $r$ around the query
image's GPS, and then randomly selected multiple \textit{reference
  tuples}. Subsequently, we used 4 different methods to estimate camera
poses with multiple \textit{reference tuples}: maximum number of
matched features (\textit{maxf}), simple average (\textit{avg}),
weighted average (\textit{wavg}) and the robust weighted average
(\textit{r-wavg}). The number of reference images was varied from
one to five.

\begin{figure*}
\begin{subfigure}{.5\textwidth}
  \centering
  \includegraphics[width=1\linewidth]{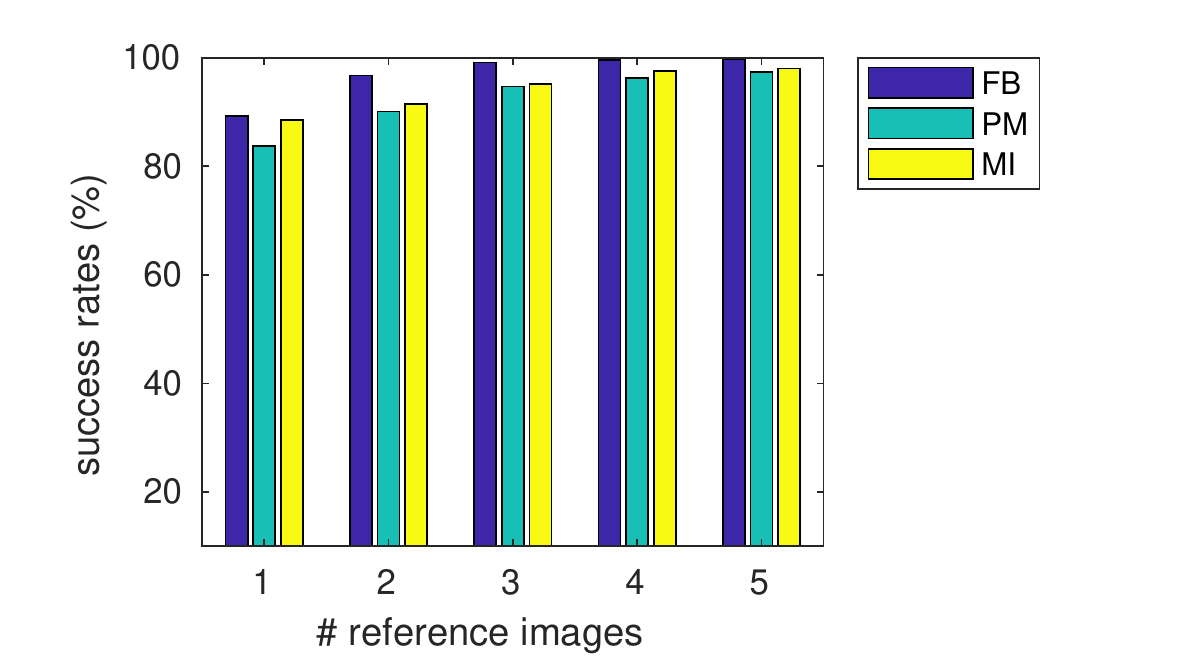}
  \caption{KITTI}
  \label{fig:exe2_kitti_rate}
\end{subfigure}%
\begin{subfigure}{.5\textwidth}
  \centering
  \includegraphics[width=1\linewidth]{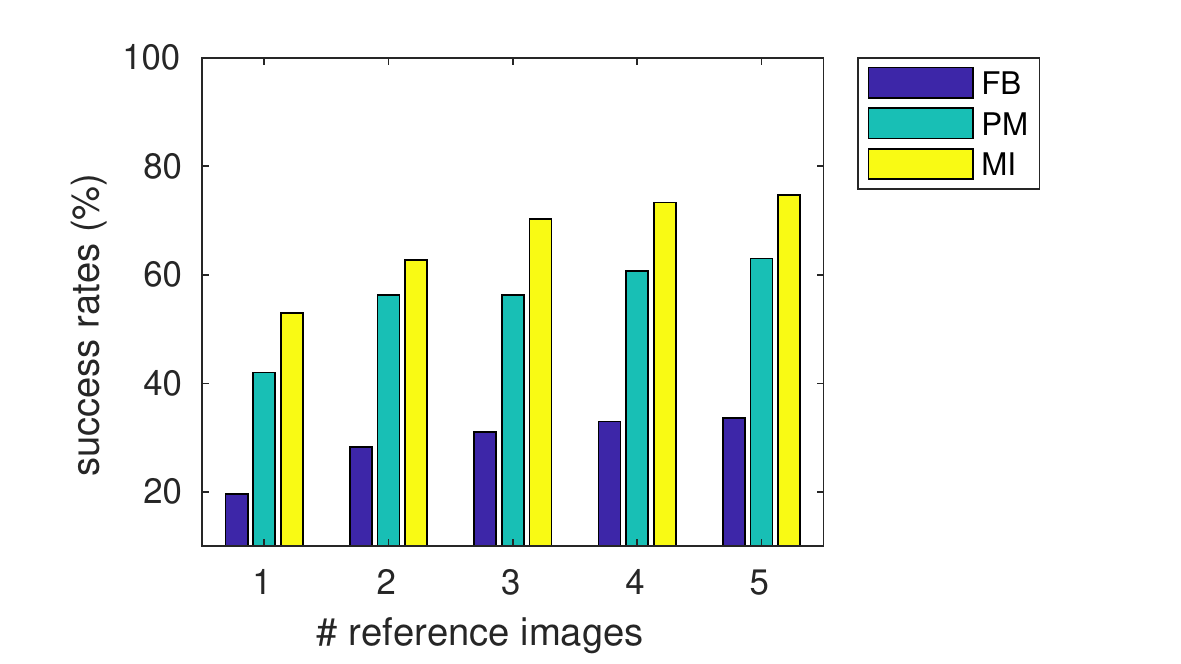}
  \caption{Oxford RobotCar}
  \label{fig:exe2_oxford_rate}
\end{subfigure}
\caption{\textbf{Success rates} comparison for three pipelines with
  multiple reference images and \textit{robust weighted average}
  merge method in two datasets. } 
\label{fig:exe2}
\end{figure*}

The comparison of different multiple-references pose estimation methods with KITTI
dataset are shown in Table~\ref{tab:kitti_exe2}.
Fig.~\ref{fig:exe2} compares the
success rates for different camera pose estimation methods with
multiple reference images  using the \textit{robust weighted average 
(r-wavg)} method in both KITTI and Oxford RobotCar datasets. 
The \textit{r-wavg} method was used
since it yielded the best overall performance for all the pose
estimation methods. 

The main findings of this experiment are that (1) the \textit{r-wavg} method 
outperforms other fusion strategies (Table~\ref{tab:kitti_exe2}), and
(2) feature-based approach is the most accurate in terms of both translation 
accuracy and orientation accuracy (Table~\ref{tab:kitti_exe2}).
Again, the results are collectively summarized in the discussion section 
(Section~\ref{sect:discussion}). 

% FB results in meres
\begin{table*}
\centering
\caption{Performance of 4 pose merge methods used by each camera pose estimation 
approach in KITTI dataset. 10\% images from one sequence are used as query image 
and the rest are used as references, and uncertainty radius is 10 meters. The reported 
results are computed from all processed images by each camera pose estimation approach.}
\resizebox{0.9\textwidth}{!}{ 
\begin{tabular}{ccc|cc|cc|cc|cc}
\hline
\#reference images & \multicolumn{2}{c|}{1} & \multicolumn{2}{c|}{2} & \multicolumn{2}{c|}{3} & \multicolumn{2}{c|}{4} & \multicolumn{2}{c}{5} \\ 
& \begin{tabular}[c]{@{}c@{}}RMSE\\ (m)\end{tabular} & \begin{tabular}[c]{@{}c@{}}RMSE\\ (deg)\end{tabular} & \begin{tabular}[c]{@{}c@{}}RMSE\\ (m)\end{tabular} & \begin{tabular}[c]{@{}c@{}}RMSE\\ (deg)\end{tabular} & \begin{tabular}[c]{@{}c@{}}RMSE\\ (m)\end{tabular} & \begin{tabular}[c]{@{}c@{}}RMSE\\ (deg)\end{tabular} & \begin{tabular}[c]{@{}c@{}}RMSE\\ (m)\end{tabular} & \begin{tabular}[c]{@{}c@{}}RMSE\\ (deg)\end{tabular} & \begin{tabular}[c]{@{}c@{}}RMSE\\ (m)\end{tabular} & \begin{tabular}[c]{@{}c@{}}RMSE\\ (deg)\end{tabular} \\
\midrule
\multicolumn{3}{c}{\em Feature-based (FB)}\\
avg & 0.125 & 1.76 & 0.216 & 2.07 & 0.248 & 2.20 & 0.212 & 1.80 & 0.195 & 1.61 \\

wavg & 0.125 & 1.76 & 0.148 & \textbf{1.67} & 0.151 & 1.78 & 0.103 & 1.22 & 0.090 & 1.11 \\

maxf & 0.125 & 1.76 & \textbf{0.106} & 1.82 & \textbf{0.093} & 1.79 & 0.060 & 1.21 & 0.049 & 1.03 \\

\textbf{r-wavg} & \textbf{0.125} & \textbf{1.76} & 0.117 & 1.70 & 0.104 & \textbf{1.59} & \textbf{0.059} & \textbf{1.13} & \textbf{0.045} & \textbf{0.93} \\
\midrule
\multicolumn{3}{c}{\em Photometric (PM)}\\
avg 	& 1.67  & 1.22 & 2.39 & 1.36 & 2.20 & 1.42 & 2.09 & 1.23 & 1.90 & 1.05 \\
wavg 	& 1.67  & 1.22 & 1.79 & 1.08 & 1.55 & 1.07 & 1.29 & 0.80 & 1.12 & 0.70  \\
maxf 	& 1.67  & 1.22 & 1.37 & 1.07 & 1.22 & 1.02 & 1.20 & 0.73 & 1.07 & 0.67  \\
\textbf{r-wavg} 	&\textbf{1.67} & \textbf{1.22} & \textbf{1.40} & \textbf{1.01} & \textbf{1.22} & \textbf{0.87} & \textbf{1.12} & \textbf{0.68} & \textbf{0.99} & \textbf{0.58}  \\
\midrule
\multicolumn{3}{c}{\em Mutual Information (MI)}\\
avg 	& 1.75  & 1.35 & 1.71 & 1.28 & 1.84 & 1.46 & 1.69 & 1.25 & 1.61 & 1.13  \\
wavg 	& 1.75  & 1.35 & 1.51 & 1.14 & 1.39 & 1.21 & 1.17 & 0.79 & 1.18 & 0.78  \\
maxf 	& 1.75  & 1.35 & 1.43 & 1.10 & 1.29 & 1.06 & 1.17 & 0.80 & 1.13 & 0.68  \\
\textbf{r-wavg} 	& \textbf{1.75}  & \textbf{1.35} & \textbf{1.43} & \textbf{1.07} & \textbf{1.26} & \textbf{0.95} & \textbf{1.10} & \textbf{0.69} & \textbf{1.03} & \textbf{0.62}  \\
\bottomrule
\end{tabular}
}

\label{tab:kitti_exe2}
\end{table*}

\subsection{Experiments at large uncertainty}
\label{subsec:exp_large_uncertainty}

Based on the strong empirical results in Section~\ref{subsec:exp_single_reference}, 
we proposed a hybrid approach that takes the advantages of both the 
feature-based and the 
mutual-information-based approaches as described in Section~\ref{sect:hybrid}. 
In this Section, we tested these
4 camera pose estimation methods (feature-based, photometric-based, mutual-information-based, and hybrid approaches) with maximum 5 reference images under large 
uncertainty condition.

In Section~\ref{sec:pose_large_uncertainties}, we introduced a framework for
camera pose estimation under large location uncertainty. In the extreme
case this means that no prior location estimate is available, but
the query image must be matched to the whole reference database, so a 
image retrieval method~\citep{philbin2007object} is applied
to find the reference images. In our
experiment we used 200 meters as the initial uncertainty radius for the KITTI
and 50 meters for the Oxford dataset, adopted the
multiple reference (up to 5 reference images) to improve robustness of
all investigated methods. The KITTI dataset correspond to an
``ideal case'' where query images and references images are from the same 
environmental setting, while Oxford RobotCar dataset represents results for a more realistic
case where the query and references images have completely different environmental 
setting. We report results for all 11 KITTI sequences, and a
100-fold experimental results for the Oxford RobotCar dataset where one sequence used as the
reference dataset and other sequences as the query dataset. 
 
The results for the 11 KITTI sequences
are shown in Table~\ref{tab:exe3_kitti}. This table consists of 44 experiments, 
each of the 11
sequence in KITTI dataset processed by the four camera pose estimation
methods. The uncertainty radius was set to be 200 meters, the
maximum number of reference images was set to 5, and we classify
the localization failure as either system-reported or $\ge$ 10 meters
absolute translation error. The
results for the Oxford RobotCar dataset are shown in
Table~\ref{tab:exe3_oxford}. The uncertainty radius was set to be 50 meters, the
maximum number of reference images was set to 5, and we classify the localization failure
as either system-reported or 20 meters absolute translation error. 

The most interesting
finding of these experiments is that our hybrid method that combines the
complementary properties of the feature-based and mutual information
based approaches is the most effective and robust for all
query-reference pairs with the difficult and realistic Oxford dataset. The detailed
analysis is presented the discussion Section~\ref{sect:discussion}.

\begin{table*}
\centering
\caption{Large uncertainty pose estimation results for 11 sequences in the KITTI dataset combining image retrieval and pose estimation. 
The uncertainty radius is 200 meters and the number of automatically retrieved reference images is 5. Note these two original 
papers~\citep{tykkala2013photorealistic, nid-slam} were designed for slam problem, but we modified the algorithms to adjust to our problem.}
\resizebox{0.9\textwidth}{!}{ 
\begin{tabular}{@{}lccc|ccc|ccc|ccc@{}}
\hline

\#sequence ID & \multicolumn{3}{c|}{00} & \multicolumn{3}{c|}{01} & \multicolumn{3}{c|}{02} & \multicolumn{3}{c}{03} \\ 
 & \% & \begin{tabular}[c]{@{}c@{}}RMSE \\ (m)\end{tabular} & \begin{tabular}[c]{@{}c@{}}RMSE\\ (deg)\end{tabular} & \% & \begin{tabular}[c]{@{}c@{}}RMSE\\ (m)\end{tabular} & \begin{tabular}[c]{@{}c@{}}RMSE\\ (deg)\end{tabular} & \% & \begin{tabular}[c]{@{}c@{}}RMSE\\ (m)\end{tabular} & \begin{tabular}[c]{@{}c@{}}RMSE\\ (deg)\end{tabular} & \% & \begin{tabular}[c]{@{}c@{}}RMSE\\ (m)\end{tabular} & \begin{tabular}[c]{@{}c@{}}RMSE\\ (deg)\end{tabular} \\

FB~\citep{2d_2d_3d} & \textbf{99.8} & 0.031 & 0.676 & 84.5 & 0.494 & 0.567 & \textbf{99.8} & 0.025 & 0.415 & \textbf{100} & 0.015 & 0.370 \\

PM~\citep{tykkala2013photorealistic} & 98.2 &  0.603 & 0.423 & 76.4 & 1.208 & 0.343 & 92.9 & 0.550 & 0.324 & 98.8 & 0.342 & 0.279 \\

MI~\citep{nid-slam} & 97.8 & 0.633 & 0.415 & 60.0 & 0.980 & 0.353 & 97.6 & 0.475 & 0.327 & 98.8 & 0.270 & 0.223 \\

HY (proposed) & \textbf{99.8} & 0.031 & 0.676 & \textbf{89.1} & 0.505 & 0.562 & \textbf{99.8} & 0.025 & 0.415 & \textbf{100} & 0.015 & 0.370 \\

 &  &  &  &  &  &  &  &  &  &  &  &  \\ \hline

\#sequence ID & \multicolumn{3}{c|}{04} & \multicolumn{3}{c|}{05} & \multicolumn{3}{c|}{06} & \multicolumn{3}{c}{07} \\ 
 & \% & \begin{tabular}[c]{@{}c@{}}RMSE\\ (m)\end{tabular} & \begin{tabular}[c]{@{}c@{}}RMSE\\ (deg)\end{tabular} & \% & \begin{tabular}[c]{@{}c@{}}RMSE\\ (m)\end{tabular} & \begin{tabular}[c]{@{}c@{}}RMSE\\ (deg)\end{tabular} & \% & \begin{tabular}[c]{@{}c@{}}RMSE\\ (m)\end{tabular} & \begin{tabular}[c]{@{}c@{}}RMSE\\ (deg)\end{tabular} & \% & \begin{tabular}[c]{@{}c@{}}RMSE\\ (m)\end{tabular} & \begin{tabular}[c]{@{}c@{}}RMSE\\ (deg)\end{tabular} \\

FB~\citep{2d_2d_3d} & \textbf{100} & 0.028 & 0.132 & \textbf{100} & 0.022 & 0.472 & \textbf{100} & 0.029 & 0.421 & \textbf{100} & 0.018 & 0.326 \\

PM~\citep{tykkala2013photorealistic} & 96.3 & 0.783 & 0.222 & 97.8 & 0.514 & 0.360 & 98.2 & 0.382 & 0.308 & 97.3 & 0.505 & 0.336 \\

MI~\citep{nid-slam} & \textbf{100} & 0.495 & 0.177 & 97.1 & 0.537 & 0.352 & 96.4 & 0.551 & 0.332 & 98.2 & 0.500 & 0.319 \\ 

HY (proposed) & \textbf{100} & 0.028 & 0.132 & \textbf{100} & 0.022 & 0.472 & \textbf{100} & 0.029 & 0.421 & \textbf{100} & 0.018 & 0.326 \\

 &  &  &  &  &  &  &  &  &  &  &  &  \\ 
\hline
\#sequence ID & \multicolumn{3}{c|}{08} & \multicolumn{3}{c|}{09} & \multicolumn{3}{c|}{10} & \multicolumn{1}{l}{} & \multicolumn{1}{l}{} & \multicolumn{1}{l}{} \\
 & \% & \begin{tabular}[c]{@{}c@{}}RMSE\\ (m)\end{tabular} & \begin{tabular}[c]{@{}c@{}}RMSE\\ (deg)\end{tabular} & \% & \begin{tabular}[c]{@{}c@{}}RMSE\\ (m)\end{tabular} & \begin{tabular}[c]{@{}c@{}}RMSE\\ (deg)\end{tabular} & \% & \begin{tabular}[c]{@{}c@{}}RMSE\\ (m)\end{tabular} & \begin{tabular}[c]{@{}c@{}}RMSE\\ (deg)\end{tabular} & \multicolumn{1}{l}{} & \multicolumn{1}{l}{} & \multicolumn{1}{l}{} \\

FB~\citep{2d_2d_3d}  & \textbf{100} & 0.018 & 0.383 & 99.4 & 0.019 & 0.356 & \textbf{100} & 0.019 & 0.420 & \multicolumn{1}{l}{} & \multicolumn{1}{l}{} & \multicolumn{1}{l}{} \\

PM~\citep{tykkala2013photorealistic} & 97.3 & 0.499 & 0.329 & 95.0 & 0.548 & 0.321 & 94.2 & 0.634 & 0.355 & \multicolumn{1}{l}{} & \multicolumn{1}{l}{} & \multicolumn{1}{l}{} \\

MI~\citep{nid-slam} & 95.3 & 0.518 & 0.341 & 93.7 & 0.400 & 0.350 & 91.7 & 0.780 & 0.343 & \multicolumn{1}{l}{} & \multicolumn{1}{l}{} & \multicolumn{1}{l}{} \\ 

HY (proposed) & \textbf{100} & 0.018 & 0.383 & \textbf{100} & 0.019 & 0.368 & \textbf{100} & 0.019 & 0.420 & \multicolumn{1}{l}{} & \multicolumn{1}{l}{} & \multicolumn{1}{l}{} \\

\hline
\end{tabular}
}
\label{tab:exe3_kitti}
\end{table*}

% succuss: (1)system success, (2) trans_err < 20m
\begin{table*}[]
\centering
\caption{Large uncertainty pose estimation results for the 5 different
  sequences in Oxford dataset (5-fold experiment where each sequence was paired
  with each sequence to form query-reference pairs).
  The uncertainty radius was set to 50 meters and the number of
  automatically retrieved reference images was 5. The failure threshold was set to 20 meters.
  Note these two original papers~\citep{tykkala2013photorealistic, nid-slam} were designed 
  for slam problem, but we modified the algorithms to adjust to our problem..} 
\resizebox{0.9\textwidth}{!}{ 
\begin{tabular}{clccc|ccc|ccc|ccc|ccl}
\hline
\multicolumn{2}{c}{\multirow{3}{*}{}} & \multicolumn{3}{c|}{overcast} & \multicolumn{3}{c|}{sun} & \multicolumn{3}{c|}{night} & \multicolumn{3}{c|}{rain} & \multicolumn{3}{c}{snow} \\
\multicolumn{2}{c}{} & \multirow{2}{*}{\%} & RMSE & RMSE & \multirow{2}{*}{\%} & RMSE & RMSE & \multirow{2}{*}{\%} & RMSE & RMSE & \multirow{2}{*}{\%} & RMSE & RMSE & \multirow{2}{*}{\%} & RMSE & \multicolumn{1}{c}{RMSE} \\
\multicolumn{2}{c}{} &  & (m) & (deg) &  & (m) & (deg) &  & (m) & (deg) &  & (m) & (deg) &  & (m) & \multicolumn{1}{c}{(deg)} \\ 
\hline
\multirow{4}{*}{overcast} & FB~\citep{2d_2d_3d} & 98.4 & 0.111 & 0.791 & 32.7 & 3.309 & 3.070 & 6.0 & 6.441 & 4.200 & 30.7 & 3.491 & 5.029 & 17.0 & 1.505 & 7.068 \\
 & PM~\citep{tykkala2013photorealistic} & 100.0 & 1.601 & 0.568 & 28.8 & 14.389 & 9.250 & 23.2 & 14.283 & 19.964 & 22.2 & 12.004 & 4.122 & 21.6 & 12.202 & 8.342 \\
 & MI~\citep{nid-slam} & 100.0 & 1.577 & 0.726 & 41.6 & 11.603 & 7.945 & 43.8 & 13.077 & 12.788 & 37.3 & 13.523 & 12.607 & 44.3 & 11.594 & 8.918 \\
 & HY (proposed) & \textbf{100.0} & 0.112 & 0.788 & \textbf{57.2} & 4.366 & 6.840 & \textbf{47.6} & 12.470 & 9.867 & \textbf{55.1} & 5.672 & 8.112 & \textbf{52.3} & 8.124 & 8.360 \\

\hline 
 
\multirow{4}{*}{sun} & FB~\citep{2d_2d_3d} & 34.0 & 2.615 & 2.122 & 98.3 & 0.121 & 0.706 & 3.3 & 5.104 & 14.612 & 13.6 & 2.982 & 4.482 & 16.6 & 2.970 & 4.820 \\
 & PM~\citep{tykkala2013photorealistic} & 30.7 & 11.482 & 6.946 & 99.7 & 2.035 & 0.597 & 31.5 & 12.904 & 10.508 & 25.3 & 12.819 & 8.778 & 22.8 & 14.221 & 5.644 \\
 & MI~\citep{nid-slam} & 44.0 & 11.908 &  6.384 & 99.7 & 1.770  & 0.567  & 40.7 & 11.246 & 7.906 & 35.0 & 12.007 & 7.063 & 40.4 & 11.754 & 4.982 \\
 & HY (proposed) & \textbf{56.7} & 3.478 & 3.983 & \textbf{99.7} & 0.122 & 0.715 & \textbf{41.8} & 10.584 & 8.810 & \textbf{44.0} & 10.225 & 5.954 & \textbf{48.7} & 9.569 & 4.955 \\

\hline 
 
\multirow{4}{*}{night} & FB~\citep{2d_2d_3d} & 5.9 & 4.132 & 3.158 & 2.5 & 5.759 & 16.003 & 91.1 & 0.221 & 0.750 & 1.2 & 2.879 & 4.171 & 2.7 & 6.481 & 8.260 \\
 & PM~\citep{tykkala2013photorealistic} & 22.5 & 12.604 & 11.445 & 16.1 & 14.336 & 17.676 & 99.3 & 2.507 & 0.566 & 16.3 & 12.824 & 11.574 & 12.7 & 13.751 & 19.656 \\
 & MI~\citep{nid-slam} & 31.4 & 12.698 & 11.425 & 34.7 & 12.035 & 7.351 & 99.0  & 2.247  & 0.580  & 37.2 & 12.382 & 4.733 & 32.7 & 11.899 & 7.023 \\
 & HY (proposed) & \textbf{36.3} & 11.334 & 8.960 & \textbf{35.8} & 12.004 & 7.584 &  \textbf{99.3} & 0.238  &  0.887 & \textbf{37.6} & 12.236 & 4.671 & \textbf{33.3} & 11.381 & 7.477 \\
 
\hline 
 
\multirow{4}{*}{rain} & FB~\citep{2d_2d_3d} & 33.9 & 3.279 & 2.908 & 14.6 & 2.330 & 5.389 & 4.0 & 2.345 & 1.505 & 97.3 & 0.192 & 0.767 & 18.5 & 3.455 & 3.844 \\
 & PM~\citep{tykkala2013photorealistic} & 31.6 & 11.097 & 4.481 & 37.6 & 12.831 & 7.472 & 25.3 & 12.732 & 6.548 & 100.0 & 2.417 & 0.610 & 29.5 & 13.204 & 5.913 \\
 & MI~\citep{nid-slam} & 34.5 & 12.853 & 9.739 & 39.0 & 10.986 & 6.528 & 33.3 & 13.332 & 8.483 & 100.0  & 1.966  & 1.077  & 39.1 & 11.141 & 6.673 \\
 & HY (proposed) & \textbf{56.1} & 4.322 & 5.119 & \textbf{46.7} & 8.891 & 7.420 & \textbf{35.9} & 11.276 & 7.240 & \textbf{100.0} & 0.202 & 0.773 & \textbf{49.0} & 6.494 & 5.707 \\

\hline 
 
\multirow{4}{*}{snow} & FB~\citep{2d_2d_3d} & 10.8 & 2.556 & 3.880 & 11.5 & 2.731 & 8.554 & 2.2 & 5.733 & 13.984 & 11.6 & 2.622 & 4.836 & 97.7 & 0.145 & 0.834\\
 & PM~\citep{tykkala2013photorealistic} & 18.9 & 12.768 & 36.815 & 16.4 & 15.801 & 29.476 & 20.5 & 14.563 & 5.459 & 12.0 & 13.549 & 10.648 & 100.0 & 2.200 & 0.559 \\
 & MI~\citep{nid-slam} & 32.4 & 11.947 & 9.698 & 35.9 & 12.708 & 7.956 & 34.8 & 12.174 & 6.299 & 27.6 & 12.120 & 6.223 &  100.0 &  2.133 &  0.731 \\
 & HY (proposed) & \textbf{36.0} & 9.348 & 7.545 & \textbf{42.9} & 10.170 & 8.480 & \textbf{35.9} & 11.849 & 6.278 & \textbf{35.6} & 9.851 & 7.310 & \textbf{100.0} & 0.149 & 0.804\\
 \hline
\end{tabular}
}

\label{tab:exe3_oxford}
\end{table*}

%%%%%%%%%%%%%%%%%%%%%%%%%%%%%%%%%%%%%%%%%%%%%%%%%%%%%%%%%%%%%%%%%%%%%%%%%%%%%%%
\section{Discussion}
\label{sect:discussion}

\subsection{Camera pose estimation with single reference image}

Table~\ref{tab:exe1} reports both translation errors and orientation errors 
comparisons for three strategies using a single \textit{reference tuple}. 
In Table~\ref{tab:exe1}, the numbers of images in the second line of each sub-table 
are the number of images that all methods successfully processed and therefore the
error numbers are comparable between the methods, and the percentages on the third 
line of each sub-table are the corresponding percentages of the
successfully processed images among the total number of images.

From Table~\ref{tab:kitti_exe1_trans} and~\ref{tab:oxford_exe1_trans}, we have two findings: 

\begin{enumerate}

\item By looking into each column, we find that as long as the
  feature-based approach is able to estimate the camera pose (minimum 
  4 consistent 2D-3D correspondences are required to compute the camera 
  pose by the PnP method~\citep{pnp_2})), its
  estimated camera poses have smaller translation errors than the
  other two methods in both KITTI and Oxford RobotCar dataset. This
  result indicates that the feature-based approach is more accurate
  in pose estimation in both ideal environment conditions
  (KITTI dataset) and realistic environment conditions (Oxford
  RobotCar dataset). 

\item By looking into each row, we find that the translation errors of
  both photometric-based and mutual-information-based approach
  increase with the increase of the \textbf{uncertainty radius}, but the
  translation errors of feature-based approach do not vary much. Since
  the closer reference images bring better initialization for both the
  photometric-based and mutual-information-based approach, the results
  indicate that both the photometric-based and mutual-information-based 
  approach are sensitive to the initialization. However, the
  feature-based approach is much less sensitive to the location of the
  reference images.

\end{enumerate}

Table~\ref{tab:kitti_exe1_ori} and~\ref{tab:oxford_exe1_ori} compare 
the orientation errors for the studied methods. Among these different 
camera pose estimation methods, the differences between their orientation 
errors are small. In other words, all these methods 
perform similarly in terms of orientation error for both KITTI and 
Oxford RobotCar datasets. The reason might be that all the images are 
taken by a camera mounted on a car driving along the street, so the 
query images and the reference images may share similar viewpoints.

Fig.~\ref{fig:exe1} records the success rate (see definitions in
Section~\ref{subsct:measures}) comparison for the three strategies with
single \textit{reference tuple} at different uncertainty ranges in two
public datasets. Fig.~\ref{fig:exe1} shows the following: 

\begin{enumerate}

\item By looking at the three bars corresponding to each uncertainty
  radius, the feature-based approach has higher success rate than the
  other two approaches in KITTI dataset; however, feature-based
  approach has the lowest success rate among all three approaches in
  Oxford RobotCar dataset. The mutual-information-based approach has
  the highest success rate in Oxford RobotCar dataset. In other words,
  the success rate of feature-based approach is greatly influenced by
  the environmental conditions between the query and reference
  images. On the other hand, the mutual-information-based approach is
  the most robust in terms of the success rate under different
  environmental conditions. 

\item When analyzing the same pose estimation method for different
  uncertainty radii, the success rates of all approaches go down with
  the increase of the uncertainty radius.  

\end{enumerate}

The state-of-the-art SLAM approach~\citep{nid-slam} claims that the
mutual-information-based SLAM approach has higher success rate than
the state-of-the-art feature-based SLAM
approach~\citep{orb-slam2}. Our experiment in
Fig.~\ref{fig:oxford_exe1_rate} provides the same conclusion under the
problem of 6-DoF camera pose estimation using single reference image
and 3D point cloud. Interestingly enough, our experiments in
Table~\ref{tab:exe1} tells that the feature-based approach can
be more accurate as long as it is able to compute the camera pose. 

%From the above experiment, we can conclude that feature-based approach is able to give accurate camera pose by using only 4 consistent matched points~\citep{pnp_2}, the mutual-information-based approach is accurate even in the challenging environment conditions. On the other hand, the mutual-information-based approach is computational more expensive but it is more robust than the feature-based approach in challenging environment conditions; in the challenging environment, if there are more than 4 consistent matched features, the feature-based approach is able to compute the reliable camera pose, but it is usually hard to find enough consistent matches.

%
%
\subsection{Camera pose estimation with multiple reference images}

Table~\ref{tab:kitti_exe2} reports the results for
the experiments with multiple reference
images for three different camera pose estimation
methods. The results show that fusing the poses from multi-reference 
improves the performance of the camera pose estimation results, and 
\textit{robust weighted average} (\textit{r-wavg}) outperforms the other approaches,
  especially with the increased number of reference images. 
  
Fig.~\ref{fig:exe2} compares the success rates of the different approaches
with multiple reference images using \textit{robust weighted
  average} method in both KITTI and Oxford RobotCar
datasets. Fig.~\ref{fig:exe2} tells us two things: 

\begin{enumerate}
\item By looking into the success rate of each method, we see that the
  success rate of different pose estimation methods increases with
  increasing number of reference images regardless of the
  environmental conditions of the query and reference images. 

\item By looking into the three bars at each plot, it shows that the
  feature-based approach has the highest success rate among different
  approaches in the KITTI dataset, but has the lowest success rate in
  Oxford RobotCar dataset. Instead, the mutual-information-based
  approach has the highest success rate for Oxford RobotCar
  dataset. In other words, mutual information is more robust than the
  two other approaches under changing environmental conditions, which
  finding is consistent with the single reference experiments.

\end{enumerate}

In the literature, the camera pose estimation usually requires
geometry verification~\citep{Sattler_2016_CVPR} which is very
effective but requires extra computation. This \textit{robust weighted 
average} method is a light approach and can be easily adapted with any pose
estimation method. 

\subsection{Camera pose estimation at large uncertainties}

By looking at the columns of
success rates in Table~\ref{tab:exe3_kitti}, we see that the hybrid and
feature-based approaches outperform other methods in cases where the query 
and reference images have been captured at similar imaging conditions 
(KITTI dataset). The hybrid approach performs similarly as
the feature-based approach which indicates that the proposed hybrid
method can retain good properties of the feature-based method. For
the sequence 01 hybrid is superior. This has the
explanation that 01 is captured from highway
(Table~\ref{tab:kitti_summary}) where there
are less reliable features
to be found than in urban scenes. In urban scenes hybrid and
feature-based methods provide practically the same accuracy.

Table~\ref{tab:exe3_oxford} summarizes the results from 100 experiments
where the four camera pose estimation methods were used in 25 query-reference
combinations of the Oxford RobotCar dataset. In addition to a large 
displacement, query and reference images have 
been acquired at very different imaging conditions. Table~\ref{tab:exe3_oxford} 
provides a confusion matrix for the experiment combining different 
imaging conditions.  By looking into the columns of success rate in 
that table, our findings are as follows: 
\begin{enumerate}
\item mutual-information-based approach is more robust than the feature-based
  or photometric-based approaches, which is consistent with the
  findings from both Fig.~\ref{fig:exe1} and Fig.~\ref{fig:exe2}. 

\item The hybrid approach outperforms all other
  approaches in success rate when the query and reference
  images have very different imaging conditions. This confirms that
  the proposed hybrid method leverages complementary properties
  of the feature-based and mutual-information-based methods.

\end{enumerate}

The results on the diagonal of Table \ref{tab:exe3_oxford} are consistent with previous experiments in the
KITTI dataset in Table~\ref{tab:exe3_kitti}, i.e. in the ideal case
when query and reference images come from the same sequence and imaging conditions. 
In this case, feature-based
and our hybrid method outperform the other approaches. A remarkable result in 
Table\ref{tab:exe3_oxford} is that, even in the worst case scenario, the lowest 
success rate of the proposed hybrid method is 35.8\%. Recent results in the same 
dataset in similar conditions have reported success rates as low as 0 \% using 
SLAM \citep{nid-slam}. Notice that the experimental settings in that work \citep{nid-slam} 
are different from ours, but this helps understanding the difficulty of the pose 
estimation problem under challenging environmental conditions.

%%%%%%%%%%%%%%%%%%%%%%%%%%%%%%%%%%%%%%%%%%%%%%%%%%%%%%%%%%%%%%%%%%%%%%%%%%%%%%%
\section{Conclusion}
\label{sect:conclusion}

We performed systematic and extensive comparisons of three different strategies in
6-DoF camera pose estimation using reference images and 3D point
cloud: an \textit{indirect} feature-based approach, a \textit{direct} photometric-based approach
and a \textit{direct} mutual-information-based approach. ``\textit{Direct}'' in this context 
means the minimization of the cost function is done directly in the space of 
6D camera pose. In our experiments the
feature-based approach is more accurate than both the photometric-based and 
mutual-information based approaches when as
few as 4 consistent feature points are found between a query and
reference image. The mutual-information-based approach is the more robust than 
the feature-based and photometric-based approaches which
means that it can provide a moderate estimate even in the cases when
the feature-based method fails. Robustness and accuracy of all methods were improved 
with increased number of reference images, and \textit{robust weighted average} method 
outperformed other fusing methods for multiple reference images. Based on the strong 
empirical results and inspired by the complementary properties of the
feature-based and mutual-information-based approaches, we
proposed a computationally cheap and easy-to-adapt hybrid approach that
combines these two methods. In all experiments, the hybrid
method is on pair or superior. This is particularly so in challenging scenarios such as 
the Oxford RobotCar dataset, where the hybrid approach outperforms feature-based and 
mutual-information-based approaches respectively
by the average of 25.1\% and 5.8\% in terms of success rate.

% 25.1 = (100-98.4 + 57.2-32.7 + 47.6-6 + 55.1-30.7 + 52.3-17 + 56.7-34 + 99.7-98.3 + 41.8-3.3 + 44-13.6 + 48.7-16.6 + 36.3-5.9 + 35.8-2.5 + 99.3-91.1 + 37.6-1.2 + 33.3-2.7 + 56.1-33.9 + 46.7-14.6 + 35.9-4 +100-97.3 + 49-18.5 +36-10.8 + 42.9-11.5 + 35.9-2.2 + 35.6-11.6 +100-97.7)/25
% 5.8 = (100-100 + 57.2-41.6 + 47.6-43.8 + 55.1-37.3 + 52.3-44.3 + 56.7-44 + 99.7-99.7 + 41.8-40.7 + 44-35 + 48.7-40.4 + 36.3-31.4 + 35.8-34.7 + 99.3-99 + 37.6-37.2 + 33.3-32.7 + 56.1-34.5 + 46.7-39 + 35.9-33.3 +100-100 + 49-39.1 +36-32.4 + 42.9-35.9 + 35.9-34.8 + 35.6-27.6 +100-100)/25

\bibliographystyle{model2-names}
%\bibliography{refs}
%\bibliography{egbib}

\clearpage
\appendix
%%%%%%%%%%%%%%%%%%%%%%%%%%%%%%%%%%%%%%%%%%%%%%%%%%%%%%%%%%%%%%%%%%%%%%%%%%%%%%%
\section{Indirect feature-based pose estimation}
\label{append:fb}
This appendix presents the detailed description of the four stages of
the \textit{indirect} feature-based pose estimation method presented in
Section~\ref{sect:feature}. 

\subsection{Feature detection and description} 
\label{subsec:feature_detect} 

The first step of the system is to detect and extract features of
salient locations in the query and reference images. Specifically, a
feature detector is used for finding the salient points of an image,
and a feature descriptor is used to describe the neighborhood
surrounding that salient point.  

Feature detectors can extract different types of image structures,
e.g. corners~\citep{rosten2006corner, mikolajczyk2004scale},
blobs~\citep{lowe1999sift,bay2006surf, kadir2001saliency} or
regions~\citep{matas2004mser, tuytelaars2000wide,
  tuytelaars2004matching, mori2004recovering}. In turn, feature
descriptors can be divided into following categories based on their
approaches: local binary descriptors~\citep{lbp,rlbp,zhao2007dynamic,
  froba2004face, calonder2010brief, rublee2011orb,
  leutenegger2011brisk, alahi2012freak}, spectral
descriptors~\citep{lowe1999sift, lienhart2002haarLike, bay2006surf,
  dalal2005HOG, tola2010daisy, ambai2011card}, basis space
descriptors~\citep{zahn1972fourier, csurka2004BOW}, polygon shape
descriptors~\citep{matas2004mser, belongie2001shape}, 3D and
volumetric descriptors~\citep{klaser20083D-HOG,
  scovanner20073DSIFT}. In the literature, many feature descriptors,
such as SURF~\citep{bay2006surf}, BRISK~\citep{leutenegger2011brisk}
and others, provide their own detector method along with the
descriptor method. DoG~\citep{lowe1999sift} and
SURF~\citep{bay2006surf} detectors were designed for efficiency and
the other properties are slightly compromised. However, for most
applications they are still more than
sufficient~\citep{tuytelaars2008local}. 

In this work we have utilized
SURF for both feature detection and description due to its speed, performance,
and widespread use in multiple applications.

%The output of feature detection is a set of feature vectors that describe a set of salient points in the image.

% ~\citep{krig2014interest}

\subsection{Feature matching}
\label{subsec:feature_matching}

Based on the previously computed feature descriptors, the aim of
feature matching is finding 2D-to-2D correspondences between feature
points in the query and reference image.  

The popular approaches for feature matching are \textit{exhaustive
  search}, \textit{hashing}~\citep{strecha2012ldahash}, and
\textit{nearest neighbor
  techniques}~\citep{friedman1977kd-tree,lowe2004distinctive,
  muja2009fastMatching}. \textit{Exhaustive search} is achieved by
minimizing pairwise distance measures between the feature vectors of
the reference and query image. The \textit{hashing} approach reduces
the size of the descriptors by finding a more compact representation,
e.g. binary strings~\citep{strecha2012ldahash}. In \textit{nearest
  neighbor techniques}, KD-trees~\citep{friedman1977kd-tree} and their
variants~\citep{lowe2004distinctive, muja2009fastMatching} are
commonly used to quickly find approximate nearest neighbors in a
relatively low-dimensional real-valued space. The algorithm works by
recursively partitioning the set of training instances based on a
median value of a chosen attribute~\citep{friedman1977kd-tree}.  

We use the exhaustive search approach and adopt a minimum Euclidean
distance on the descriptor vector. For each feature point in one image,
we find the nearest neighbor as its corresponding feature point in the
other image. Besides, we reject some ambiguous matches by comparing
the distance of the closest neighbor to that of the second-closest
neighbor. In other words, correct matches need to have the closest
neighbor significantly closer than the second closest match to
achieve reliable matching~\citep{lowe2004distinctive}. The output of
the feature matching steps are a set $c$ of $n$ 2D-to-2D
correspondences between the query image $I_Q$ and reference image $I_R$:  
\begin{equation}
c = \{ (\textbf{p}_{Q}^{(1)}, \textbf{p}_{R}^{(1)}), (\textbf{p}_{Q}^{(2)}, 
\textbf{p}_{R}^{(2)}), \ldots,(\textbf{p}_{Q}^{(n)}, \textbf{p}_{R}^{(n)}) \}
\label{eq:set_2D_3D}
\end{equation}
%
%where $\textbf{p}_{Q}^{(i)} = [u_Q^{(i)}, v_Q^{(i)}, 1]^{T}$ and $\textbf{p}_{R}^{(i)} = [u_R^{(i)}, v_R^{(i)}, 1]^{T}$  
%are the $i$th 2D feature locations on reference and query images in homogeneous coordinates. 

where $\textbf{p}_{Q}^{(i)} = [u_Q^{(i)}, v_Q^{(i)}]^{T}$ and $\textbf{p}_{R}^{(i)} = [u_R^{(i)}, v_R^{(i)}]^{T}$  
are the $i$th 2D feature locations on reference and query images.

\subsection{2D-3D correspondences}
\label{subsec:indirect_2D-3D}

\begin{figure}
	\begin{center}
		\includegraphics[width=0.7\linewidth]{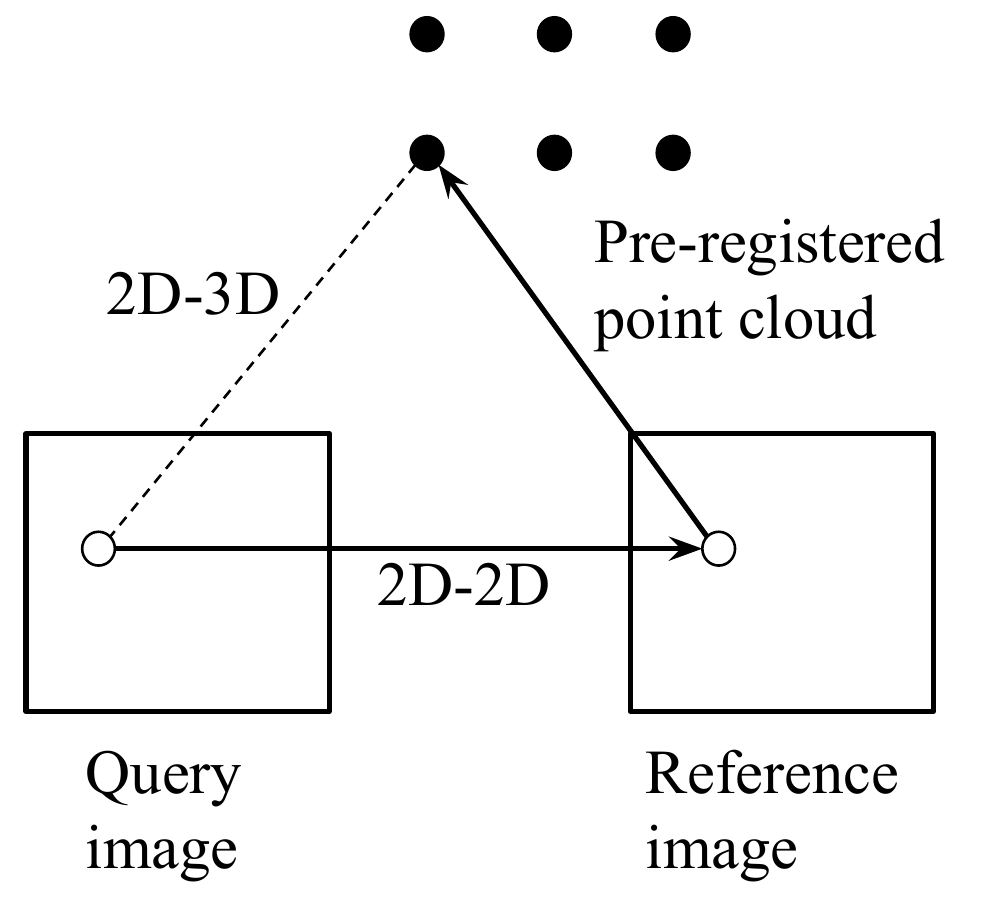}
	\end{center}
	\caption{Build 2D-3D correspondences though the 2D-2D matched features and the pre-registered point cloud.}
	\label{fig:2D_3D}
\end{figure}

The 2D-3D correspondences between the query image and the 3D point cloud are
established by using the set $c$ of 2D-2D matches and the pre-registered point cloud $P_R$.
Since the point cloud $P_R$ and the reference image $I_R$ are pre-registered and defined in the same
world coordinate system, with the 2D-2D matched features, we could indirectly link the 
2D-3D correspondences as illustrated in Fig.~\ref{fig:2D_3D}.   

However, if the matched 2D features at the reference image do not have associated 3D points from 
the pre-registered point cloud, we need to compute the 2D-3D correspondences by following steps: 
(1) project 3D point cloud 
onto the reference image, (2) compute the depth of the feature points, (3) find the corresponding 3D coordinates.

Firstly, we project the 3D point cloud $P_R = [\textbf{P}_{R}^{(1)}, \textbf{P}_{R}^{(2)}, \ldots,
\textbf{P}_{R}^{(m)}]$ onto the reference image plane, and get a set of 2D projections 
$p = [\textbf{p}^{(1)}, \textbf{p}^{(2)}, \ldots,
\textbf{p}^{(m)}]$, as shown in Fig.~\ref{fig:projections}.
For the $i$-th 3D point, $\textbf{P}_R^{(i)} = [x^{(i)}, y^{(i)}, z^{(i)}, 1]^{T}$, we
generate a 2D projection $\textbf{p}^{(i)} = [u^{(i)}, v^{(i)}]^{T}$ on the reference image plane by: 

\begin{equation}
 \textbf{p}^{(i)} = \textbf{K} \quad \textbf{M} \quad \textbf{P}_R^{(i)}
\label{eq:projection_start}
\end{equation}
where $\textbf{M}$ is the world to camera transformation matrix and $\textbf{K}$ is the intrinsic 
matrix of the reference image. $\textbf{M}$ and \textbf{K}
can be represented by~\eqref{eq:M} and~\eqref{eq:K}:

\begin{equation}
\textbf{M} = 
\left[{\begin{array}{ccc} \textbf{R} & \vert & \textbf{t}
\end{array}}\right]
\label{eq:M}
\end{equation}
where \textbf{R} is a $3\times3$ rotation matrix, and \textbf{t} is a $3\times1$ 
translation vector.

\begin{equation}
\textbf{K} = 
\left[{\begin{array}{ccc} {f_x} & \gamma & u_0 \\
0 & {f_y} & v_0 \\
0 & 0 & 1
\end{array}}\right]
\label{eq:K}
\end{equation}
where $f_x$ and $f_y$ are focal length in terms of pixels along x and y axis directions; $\gamma$ 
represents the skew coefficient between x and y axis and it is often 0; $u_0$ and $v_0$ represents 
the principle point which would ideally be in the center of the image. In the experiments of this 
paper, we assume the query image and the reference images share the camera intrinsic matrix, because 
the images from each dataset are captured with the same camera device.

%Since $\bar{\textbf{p}}_R^{(i)}$ is in homogeneous coordinates, we
%convert them to image coordinates to obtain
%$\hat{\textbf{p}}_R^{(i)}$. Subsequently, by
%applying~\eqref{eq:projection_start} on each point of $P_R$, a set
%$\hat{P}_R$ of $m$ 2D projections on the reference image plane is 
%obtained: $\hat{P}_R = \{\hat{\textbf{p}}_R^{(1)}, \hat{\textbf{p}}_R^{(2)} \ldots, \hat{\textbf{p}}_R^{(m)}\}$.

Secondly, we use nearest-neighbor search~\citep{friedman1977kd-tree} to find the closest point among 
2D projections $p$ for each 2D feature point in $c$ at the reference image.
In particular, the $j$-th feature
point $\textbf{p}_{R}^{(j)}$ in the reference image, is associated to
the $k$-th point of the 2D projection set $p$ by: 

\begin{equation}
k = NN(\textbf{p}_R^{(j)}, p), \quad k\in \{1,2\ldots,m\}
\end{equation}

Finally, we find the 3D coordinates for each 2D feature point. In particular, 
the $k$-th depth value corresponding to $\textbf{p}^{(k)}$,
namely $z^{(k)}$, is then used to find the 3D coordinates in the
reference image frame corresponding to $\textbf{p}_{R}^{(j)}$ as: 

%\begin{equation}
%\hat{\textbf{P}}^{(j)} = \textbf{K}^{-1} \Bigg[ \begin{tabular}{c} $\textbf{p}_R^{(j)}$ \\ \\ $z^{(k)}$   \end{tabular} \Bigg]
%\end{equation}
\begin{equation}
\textbf{P}^{(j)}=\begin{bmatrix}
\textbf{K}^{-1}\textbf{p}_R^{(j)} z^{(k)} \\z^{(k)}
\end{bmatrix}
\end{equation}

As a result, the final 2D-to-3D correspondences can be expressed as:

\begin{equation}
\hat{c} = \{ (\textbf{p}_{Q}^{(1)}, \textbf{P}^{(1)}), (\textbf{p}_{Q}^{(2)}, \textbf{P}^{(2)}) ..., (\textbf{p}_{Q}^{(m)}, \textbf{P}^{(m)}) \}
\end{equation}
where $ \textbf{p}_{Q}^{(i)} $ is the $i$-th 2D feature location in the query image, and $\textbf{P}^{(i)}$ is the $i$-th corresponding 3D location in the reference image coordinate.

% a kd-tree is used for the matching between the projected LIDAR points and reference image feature

% tested with both nearest neighbor and interpolation

\begin{figure}
	\begin{subfigure}{.15\textwidth}
		\centering
		\includegraphics[width=1\linewidth]{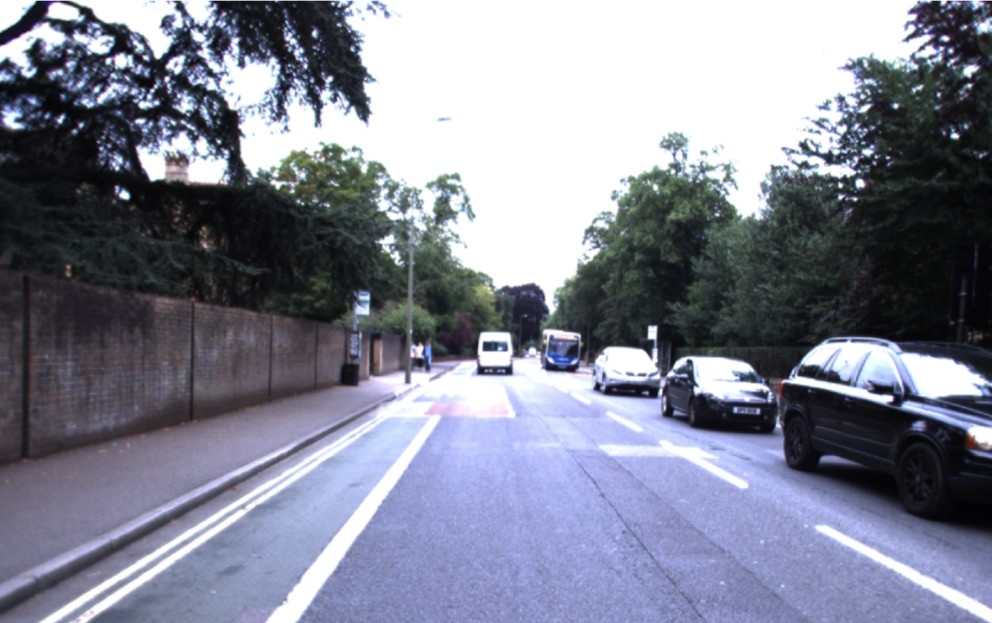}
		\caption{reference image}
		\label{fig:referenceImage}
	\end{subfigure}%
	\begin{subfigure}{.15\textwidth}
		\centering
		\includegraphics[width=1\linewidth]{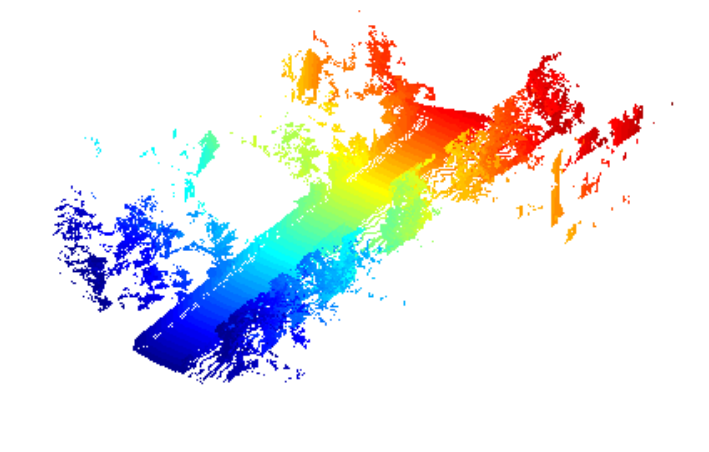}
		\caption{3D point cloud}
		\label{fig:pointCloud}
	\end{subfigure}
	\begin{subfigure}{.15\textwidth}
		\centering
		\includegraphics[width=1\linewidth]{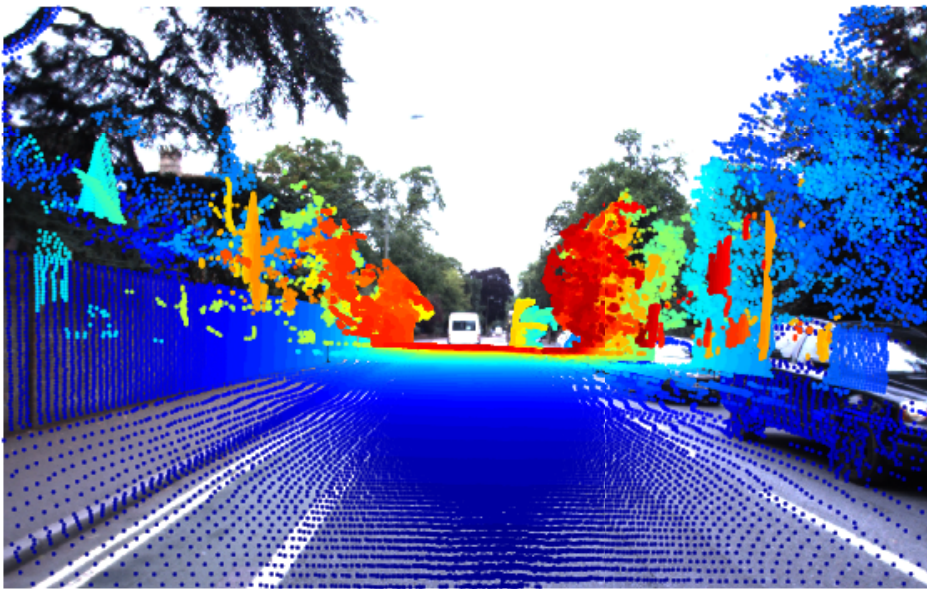}
		\caption{projections}
		\label{fig:projections}
	\end{subfigure}
	\caption{An example of projecting the 3D point cloud into the reference image.}
	\label{fig:projections}
\end{figure}

\subsection{Perspective-n-Point and RANSAC}
\label{subsec:PnP}
The set of 2D-3D correspondences $ \hat{c} $ establishes one-to-one
correspondences between 2D points in query image frame $
\textbf{p}_{Q}^{(j)} $, and 3D points in the reference image frame
$\textbf{P}^{(j)}$, for $j=1,2,\ldots,m$. The last step is to
apply Perspective-n-Point solver~\citep{pnp_2} to compute the
relative 6-DoF camera pose $\textbf{M}$ between the query image and
the reference image. For this purpose, two approaches are combined to
solve the problem: the algebraic approach and the geometric
approach. In the algebraic approach, we use Wu's zero decomposition
method to find a complete triangular decomposition of a practical
configuration for the P3P problem~\citep{pnp_2}. We can obtain up to 4
solutions for the pose using 3 points, and in the geometric approach,
we choose the solution that results in smallest squared re-projection
error for the 4th point~\citep{pnp_2},  
\begin{equation}
\textbf{M}^* = \arg \min_\textbf{M} \sum_{\forall{i}} \| \textbf{p}_{Q}^{(i)} - \textbf{K} \textbf{M} \textbf{P}^{(i)} \|, \quad	 i\in\{1,2\ldots, m\}
\end{equation}
where $ \textbf{M} $ is the sought word-to-camera transformation matrix, $ \textbf{M}^* $ is its best estimate, $\textbf{K}$ is the intrinsic matrix, $\textbf{p}_{Q}^{(i)}$ is the $i$-th feature point at the query image and $\textbf{P}^{(i)}$ is its corresponding 3D coordinate. 

In reality, the set of 2D-3D correspondences $ \hat{c} $ can be corrupted by outliers,
so it is common to use a robust estimator together with PnP slovers. RANSAC~\citep{ransac} 
estimator is a popular choice, and in our work we use a generalization of the RANSAC 
estimator, MLESAC~\citep{pnp_1}. MLESAC adopts the same sampling strategy as RANSAC to 
generate putative solutions, but chooses the solutions by maximizing the likelihood rather than just the number of inliers.

Finally, the 6-DoF camera pose can be obtained by means of the decomposition of $\textbf{M}^*$ via~\eqref{eq:M}.   

%\begin{equation}
%\textbf{M}^* = [ \textbf{R} | \textbf{t} ]
%\label{eq:getRT}
%\end{equation}
%where \textbf{R} is the rotation matrix and $\textbf{t}$ is the translation vector. 

%%%%%%%%%%%%%%%%%%%%%%%%%%%%%%%%%%%%%%%%%%%%%%%%%%%%%%%%%%%%%%%%%%%%%%%%%%%%%%%
\section{Direct photometric-based camera pose estimation}
\label{append:pm}
This appendix explains the details of the three stages of
the \textit{direct} photometric-based camera pose estimation, namely, generation of
synthetic views, direct photometric matching and coarse-to-fine
search. 

\subsection{Generation of synthetic views}
\label{subsubsec:synthetic}

The reference 3D point cloud $P_R$ does not have any color or
intensity information, but this information can be retrieved from the
reference image as follows. Firstly, we project 3D point clouds 
$P_R = [\textbf{P}_{R}^{(1)}, \textbf{P}_{R}^{(2)}, \ldots, \textbf{P}_{R}^{(m)}]$ 
onto the reference image plane using \eqref{eq:projection_start} and get a
set of 2D projections, $p = [\textbf{p}^{(1)}, \textbf{p}^{(2)}, \ldots,
\textbf{p}^{(m)}]$. This process is the same as Fig.~\ref{fig:projections}. 
Secondly, we use cubic interpolation to
compute the intensity values for each 2D projection and assign the intensity 
values to the 3D point cloud as: 
\begin{equation}
I(\textbf{P}_R^{(i)}) \leftarrow f( \textbf{p}_R^{(i)}, I_R), \quad I_R\in \mathbb{R}^2 
\end{equation}
where $I_R$ is the reference image, $\textbf{p}^{(i)}$ is the $i$-th 2D projection, 
$I(\textbf{P}_R^{(i)})$ is the intensity value of the 3D point $\textbf{P}_R^{(i)}$, 
and $f$ is the cubic interpolation function. As a result,
we assign intensity (or color) information to the 3D point cloud $P_R$.

Synthetic views can now be rendered by projecting the colored 3D point cloud
using a transformation matrix $\textbf{M}$ using \eqref{eq:projection_start}, 
and the intensities of the synthetic view $I_S$ can be obtained as:
\begin{equation}
I_S(\textbf{K} \textbf{M} \textbf{P}_R^{(i)}) \leftarrow I(\textbf{P}_R^{(i)}),
\label{eq:I_syn}
\end{equation}
where $I(\textbf{P}_R^{(i)})$ is the intensity value of the $i$-th 3D point
$\textbf{P}_R^{(i)}$, $\textbf{K}$ is the intrinsic matrix,
$\textbf{M}$ is the world-to-synthetic-view transformation, and
$I_S(\textbf{K} \textbf{M} \textbf{P}_R^{(i)})$ is the intensity value
of the projection of the 3D point $\textbf{P}_R^{(i)}$ at the
synthetic frame. Synthetic views are quickly rendered by
the standard computer graphics procedure of surface
splatting~\citep{Zwicker_SIGGRAPH-2001}. 

%\begin{figure}
%\begin{center}
%% \fbox{\rule{0pt}{2in} \rule{0.9\linewidth}{0pt}}
%   \includegraphics[width=0.8\linewidth]{source/color_lidar_horizontal.png}
%\end{center}
%   \caption{Example of the LIDAR–Street View fusion described in Section 3.1: (a) "bird-view" of the raw 3D LIDAR points; (b) "street-view" of the raw 3D LIDAR points; (c) a street view image; (d) a synthetic view of the colored LIDAR point cloud.}
%\label{fig:synthetic_image}
%\end{figure}

\subsection{Direct photometric matching}
\label{subsubsec:direct}

The \textit{direct} photometric-based approach~\citep{tykkala2013photorealistic} is defined as a 
direct minimization of the cost function at the space of 6D camera pose, and in the cost function 
it compares the pixel intensities of the query image $I_Q$ and rendered synthetic view $I_S$ from 
the colored 3D point cloud~\citep{tykkala2013photorealistic}. The task is to find the best relative 
camera transform $\textbf{M}^{*}$ that minimizes the photometric-error between query image $I_{Q}$ and
synthetic image $I_{S}$: 

\begin{equation}
\textbf{M}^{*} = \arg \min_\textbf{M} \text{RSE}(I_{Q},I_{S}),
\label{eq:rmse1}
\end{equation}

where,
\begin{equation}
\text{RSE}(I_{Q},I_{S}) = \dfrac{1}{\mu}\sum_{(u,v)\in{I_{S}}} (I_{Q}(u,v) - I_{S}(u,v))^2
\label{eq:rmse2}
\end{equation}

In~\eqref{eq:rmse2} the synthetic view $I_{S}$ is generated by \eqref{eq:I_syn}, and $\mu$ is the number of pixels in $I_{S}$. 

To improve the robustness of the matching process, we smooth the query
image $I_{S}$ by using a Gaussian filter and then we use the smoothed
version of query image in the image matching process. Moreover, 
we use M-estimator to improve the matching process, since the
M-estimator can be used for managing outliers when the residual vector
is of sufficient length for statistical
purpose~\citep{huber2011robust}. 
%Essentially an M-estimator is a function for producing uncertainty based weights for residual elements. 
The main idea is to generate small weights for residual
elements that are classified as outliers by analyzing the distribution
of residual values. Inliers always have small residual values whereas
outliers may have any error value. In our work, we
use the median filter to find the median value among the residuals,
$\mbox{RSE}(I_{Q},I_{S})$, then give zero weights to all the residual
values that are greater than the median value, and give normalized
weights to the the remain residuals. 

%way to find the spurious outliers is to use the Median Filter to find the median value among the residuals, and then give zero weights to all the residual values which are greater than the median value. Instead of the median threshold, M-estimators have some other variances. One method to determine the residual weights is by the Tukey weighting function as \eqref{eq:m-est1} to \eqref{eq:m-est3}:
%

%\begin{equation}
%U = \dfrac{\vert E \vert}{c \cdot median(\vert E \vert)}
%\label{eq:m-est2}
%\end{equation}
%

%%
%where $c=1.4826$ is the robust standard deviation, $b=4.6851$ is the Tukey specific constant, and $W(u,v)$ is the weight for the residual value, $E(u,v)$. 

With the M-estimator, we can rewrite \eqref{eq:rmse2} as the average of the weighted sum-of-square difference:
\begin{equation}
\mbox{RSE}(I_{Q},I_{S}) = \dfrac{1}{\lambda} \sum_{\forall{(u,v)}} {( E(u,v) )}^2 w(u,v)
\label{eq:rmse3}
\end{equation}
where we apply the weights to the residual vector and compute the average of the weighted sum-of-square difference, and $\lambda$ is the number of nonzero weights. $E(u,v)$ and $w(u,v)$ are defined in \eqref{eq:m-est1} and \eqref{eq:weights} as follows:

\begin{equation}
E(u,v) = (I_Q(u,v) - I_S(u,v))^2, (u,v)\in{I_{S}}
\label{eq:m-est1}
\end{equation}
where $I_Q$ is the query image, $I_S$ is the synthetic image, and $E$ is the difference between the two images.

\begin{equation}
    w(u,v)= 
\begin{cases}
    0,       & \text{if } E(u,v) > \theta\\
    1,       & \text{otherwise}
\end{cases}
\label{eq:weights}
\end{equation}
where $\theta$ is the median value of of $E(u,v)$ and $(u,v)\in{I_{S}}$.

%\begin{equation}
%W(u,v) = \begin{dcases} \left( 1-\left( \dfrac{U(u,v)}{b}\right)^{2} \right) ^{2}, \quad if \vert U(u,v) \vert \leqslant b \\ \quad  0, \quad \quad \quad \quad \quad \quad if \vert U(u,v) \vert > b   \end{dcases}
%\label{eq:m-est3}
%\end{equation}

\subsection{Coarse-to-fine grid search}
\label{subsubsec:search}

\begin{figure}
	\begin{center}
		\includegraphics[width=1\linewidth]{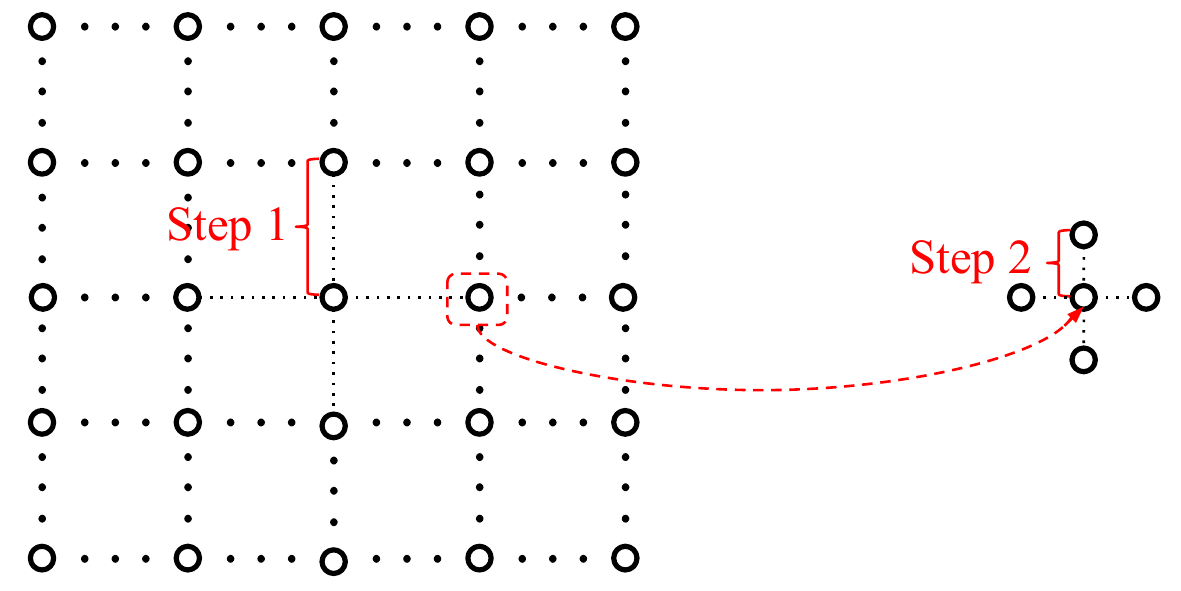}
	\end{center}
	\caption{Coarse-to-fine grid search for translation. Grids are placed along x (toward to the right of the camera) 
	and y (toward to the front of the camera) axis. Start to search the minimum with a big step (step 1) in a grid 
	manner, then follow a smaller step (step 2) search in a grid manner again at the previous minimum location.}
	\label{fig:coarse-to-fine}
\end{figure}

\begin{figure}
	\begin{center}
		\includegraphics[width=1\linewidth]{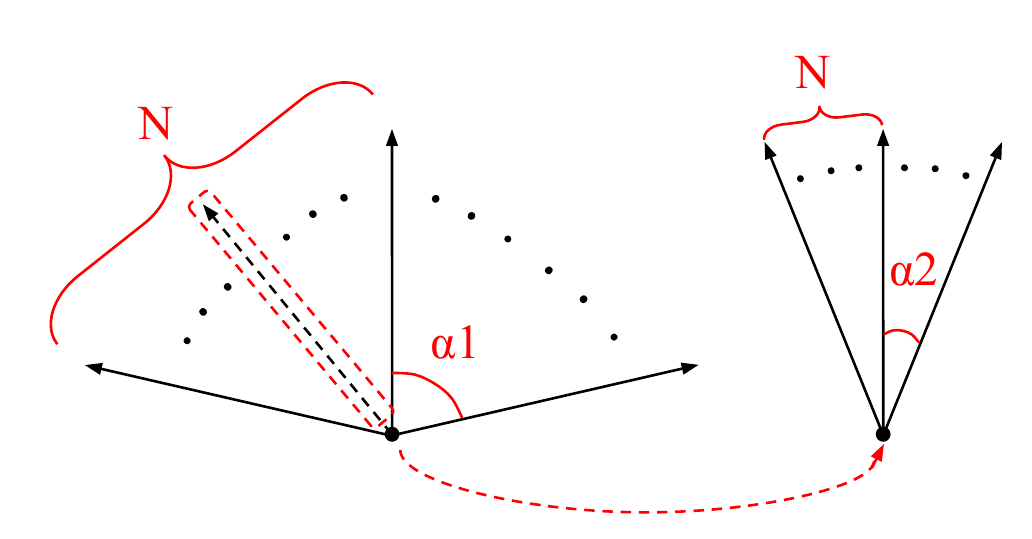}
	\end{center}
	\caption{Coarse-to-fine grid search for orientation. For the selected axis (z axis, toward up of the camera), 
	start to search the minimum with a big orientation search range $2\times \alpha 1$ with $2\times N-1$ steps, then 
	follow a smaller orientation search range $2\times \alpha 2$ with $2\times N-1$ steps in a grid manner again at 
	the previous minimum orientation.}
	\label{fig:coarse-to-fine_ori}
\end{figure}

we use a two-step coarse-to-fine grid search to solve for the matrix $\textbf{M}^*$ in \eqref{eq:rmse1},
The coarse-to-fine grid search concatenates search with a coarse step for the local
minimum with a subsequent search with at a finer step at the location
of the previous minimum location. We apply
the coarse-to-fine search firstly to the translation, and based on the 
previous minimum, we then apply it to the orientation. The process of 
the coarse-to-fine grid search is illustrated in Fig.~\ref{fig:coarse-to-fine} 
and Fig.~\ref{fig:coarse-to-fine_ori}.

Firstly, we defined a 2D search grid along x (towards the right of the camera) and z 
(toward the front of the camera) axis directions, and the origin is in the middle of the 
2D search grid. We start to search the minimum with a big step (step 1) in a grid manner, 
then follow a smaller step (step 2) search in a grid manner again at the previous minimum 
location. Fig.~\ref{fig:coarse-to-fine} illustrated the coarse-to-fine 
search for translation.

Secondly, based on the previous minimum location, we further apply the coarse-to-fine 
grid search for orientation. We could search the optimal orientations along one more 
multiple axis. For our experiments, we search the optimal orientations along the z axis 
(toward up direction of the car). As shown in Fig.~\ref{fig:coarse-to-fine_ori}, 
We start to search the minimum with a big orientation 
search range $2\times \alpha 1$ with $2\times N-1$ steps, then follow a smaller orientation 
search range $2\times \alpha 2$ with $2\times N-1$ steps in a grid manner again at the 
previous minimum orientation.

%A global coarse-to-fine search is implemented by altering the transformation
%parameters of $\prescript{synthetic}{ref}{\textbf{M}}$ (translation, orientation
%and scale) to find the minimum in Equation~\ref{eq:rmse1}, and the processes are
%as sketched in Algorithm~\ref{alg:search}.

%\begin{algorithm}[h]
%    \caption{coarse-to-fine search}
%    \label{alg:search}
%  \begin{algorithmic}[1]
%    \REQUIRE $\prescript{ref}{}{P}_{rgb}$, $K$, $I_q$, $\textbf{M}(x)\in \mathbb{SE}(3)$ where $x \in \mathbb{R}^{6}$, $Threshold$, $steps$ and $MaxItr$ 
%    \RETURN Estimated $\textbf{M}_{best}$, which minimizes the Equation~\ref{eq:rmse1}
%    \STATE \textbf{Initialization} $\textbf{M} = \textbf{Identity}$, $x=\textbf{0}$, $i=0$, $MAD_{pre} = inf$
%    \WHILE{$PercentOfValidPoints \geq Threshold$ and $i < MaxItr$}    
%	  \STATE $i = i + 1$	  
%	  \STATE $x' = x + steps, ~~steps\in \mathbb{R}^{6}$   
%      \STATE compute $I_s$ based on the current $\textbf{M}(x')\in \mathbb{SE}(3)$ via Equation~\ref{eq:I_syn}.
%      \STATE compute $MAD_{cur}$ via Equation~\ref{eq:rmse2}.      
%	  \STATE \textbf{if}~~ {$MAD_{cur} < MAD_{pre}$}
%	  \STATE ~~~~$\textbf{M}_{best} = \textbf{M}(x')$
%	  \STATE ~~~~$MAD_{pre} = MAD_{cur}$
%      \STATE \textbf{end if}
%    \ENDWHILE
%    %\Shttps://preview.overleaf.com/public/chsfghpxjkpn/images/6f24198e6b89f02567a0043bc067daee470785e3.jpegTATE return $\textbf{M}_{best}$
%  \end{algorithmic}
%\end{algorithm}

\end{document}